\pdfoutput=1
\documentclass[11pt]{article}

\usepackage{ACL2023}

\usepackage{times}
\usepackage{latexsym}

\usepackage[T1]{fontenc}
\usepackage[utf8]{inputenc}

\usepackage{microtype}

\usepackage{inconsolata}
\usepackage{amsmath}
\usepackage{xcolor}
\usepackage{mdframed}
\usepackage[skins]{tcolorbox}
\usepackage{booktabs}
\usepackage{multirow}
\usepackage{graphicx}  %%  图片包
\usepackage{subfig}    %%  子图包
\usepackage{stfloats}

\title{Deja vu: Contrastive Historical Modeling with Prefix-tuning for \\ Temporal Knowledge Graph Reasoning}

\newcommand*{\affaddr}[1]{#1} % No op here. Customize it for different styles.
\newcommand*{\affmark}[1][*]{\textsuperscript{#1}}
\newcommand*{\email}[1]{\texttt{#1}}

\author{
Miao Peng\affmark[1]$^{*}$, Ben Liu\affmark[1]\thanks{~~Equal contribution.}, Wenjie Xu\affmark[1], Zihao Jiang\affmark[1], Jiahui Zhu\affmark[2], Min Peng\affmark[1]\thanks{~~Corresponding author}\\
\affaddr{\affmark[1]School of Computer Science, Wuhan University, China}\\
\affaddr{\affmark[2]Xiaomi Inc., China}\\
\email{\{pengmiao,liuben123,vingerxu,jiangzihao,pengm\}@whu.edu.cn}\\
\email{zhujiahui1@xiaomi.com}
}

\begin{document}
\maketitle
\begin{abstract}
Temporal Knowledge Graph Reasoning (TKGR) is the task of inferring missing facts for incomplete TKGs in complex scenarios (e.g., transductive and inductive settings), which has been gaining increasing attention. Recently, to mitigate dependence on structured connections in TKGs, text-based methods have been developed to utilize rich linguistic information from entity descriptions. However, suffering from the enormous parameters and inflexibility of pre-trained language models, existing text-based methods struggle to balance the textual knowledge and temporal information with computationally expensive purpose-built training strategies. To tap the potential of text-based models for TKGR in various complex scenarios, we propose \textbf{ChapTER}, a \textbf{C}ontrastive \textbf{h}istoric\textbf{a}l modeling framework with \textbf{p}refix-tuning for \textbf{TE}mporal \textbf{R}easoning. ChapTER feeds history-contextualized text into the pseudo-Siamese encoders to strike a textual-temporal balance via contrastive estimation between queries and candidates. By introducing virtual time prefix tokens, it applies a prefix-based tuning method to facilitate the frozen PLM capable for TKGR tasks under different settings. We evaluate ChapTER on four transductive and three few-shot inductive TKGR benchmarks, and experimental results demonstrate that ChapTER achieves superior performance compared to competitive baselines with only 0.17\% tuned parameters. We conduct thorough analysis to verify the effectiveness, flexibility and efficiency of ChapTER.
\end{abstract}

\section{Introduction}

Knowledge Graphs (KGs) constitute structured representations of knowledge, storing substantial factual information in the form of (\textit{subject}, \textit{prediction}, \textit{object}). KGs have been an essential component of various NLP applications including question answering~\cite{QA-GNN}, recommendation~\cite{KGCL}, etc. Considering facts inherently evolve in KGs over time, Temporal Knowledge Graphs (TKGs) are constructed to describe the relationship between entities over time in the form of quadruple (\textit{subject}, \textit{prediction}, \textit{object}, \textit{timestamp}). While TKGs are usually incomplete, TKG reasoning (TKGR) aims to predict the missing facts from known ones. In this paper, we focus on the extrapolation task, which requires forecasting future events on TKGs with historical events. For instance, TKGR needs to answer the query (\textit{Olivia Rodrigo}, \textit{Release an album}, \textit{?}, \textit{2023-9-8}) by matching and selecting from all candidate entities based on related historical events.

To address the problem of TKG reasoning, many efforts have been made to capture temporal evolutional information in TKGs. Due to the graph-like features of TKGs, previous methods~\cite{CE-NET, CyGNet, RE-GCN} work on designing the temporal-aware encoders referring to known history and mine evolutional patterns from query neighborhoods.
Recently, pre-trained language models (PLMs) have been showing great abilities to model textual linguistic semantics, and some methods incorporate temporal information of TKGs into PLMs by designing manufactural prompts with fact texts~\cite{PPT}, tuning PLMs with a time-specific masking strategy~\cite{ECOLA}, etc. Nevertheless, on the one hand, these manually designed prompts are explicitly based on a priori assumption. On the other hand, training an entire language model on the time-specific masking strategy is computationally expensive. Though PLMs are equipped with strong linguistic inherence from the pre-training stage, they tend to exceedingly focus on the textual semantics, thus struggling to balance time-specific information and textual knowledge in TKGs.
Furthermore, given the highly dynamic essence of TKG, the continuous emergence of new unseen entities usually leads to the need for TKGR to predict entities in a more complex scenario (e.g. few-shot inductive scenario). Thus, in this paper, we focus on the research question: \textit{Can we efficiently integrate temporal history into textual knowledge in a unified PLM-based framework and adapt to TKG reasoning tasks in various complex scenarios?}

To this end, we propose ChapTER, an efficient temporal-aware PLM-based pseudo-Siamese framework adaptable for TKGR in different scenarios. Specifically, ChapTER first verbalizes the input queries and candidates with up-to-date historical contexts, and feeds them into the two-tower model encoders individually to learn history-contextualized embeddings in a decoupled manner. Then contrastive estimation is performed between them to strike a balance of temporal information and textual knowledge in representations. Rather than training an entire model, ChapTER applies a two-tower prefix-based tuning method to enable frozen PLMs capable of performing TKG reasoning tasks under both transductive and few-shot inductive settings. To refrain from the dependency of entity-related prompts in previous method~\cite{CS-PromKG}, we feed entity-agnostic virtual time prefix prompts into the frozen PLMs, empowering the model to TKGs with unseen entities.

To summarize, our contributions are as follows:
\begin{itemize}
    \item We explore a unified PLM-based pseudo-Siamese framework that can be efficiently adapted to TKGR tasks in various complex scenarios by utilizing computationally efficient prefix-based tuning.
    \item ChapTER models the historical contextual information through contrastive learning by enforcing query and candidate with highly correlated history closer and vice versa, which strikes a good balance of temporal information and textual knowledge.
    \item We evaluate ChapTER on both transductive and few-shot inductive TKGR tasks and experimental results on seven datasets demonstrate ChapTER achieves competitive performance with less than 0.17\% tuned parameters.
\end{itemize}

\section{Related Work}

\noindent \textbf{Transductive TKG Reasoning.}
Most existing transductive TKG reasoning methods are embedding-based and some of them extended from previous KG reasoning methods. TTransE~\cite{TTransE} extends the distance-based method TransE~\cite{TransE} by incorporating extra temporal constraints among facts. TNTComplEx~\cite{TNTComplEx} extends ComplEx~\cite{ComplEx} by performing 4th-order tensor factorization to learn time-aware representations. Besides, graph-based methods~\cite{RE-NET, RE-GCN, CEN} employ GCNs to capture the structural information via message passing and model temporal correlations from knowledge graph snapshots with historical information. More recently, PLM-based models have been utilized to incorporate external textual semantics for TKG reasoning. PPT~\cite{PPT} converts TKGR task into a masked token prediction task by utilizing PLM with manually designed prompts. ECOLA~\cite{ECOLA} learns the contextualized representations by jointly optimizing the TKGR and the masked language modeling objectives.

\noindent \textbf{Inductive TKG Reasoning.}
TKG reasoning tasks under inductive setting aim to predict new emerging entities in TKGs, indicating that unseen entities in the test set are not contained in the train set. To handle unseen entities, GNN-based methods like GraIL~\cite{GraIL} and NOODLE~\cite{NOODLE} extract enclosing subgraphs and learn entity-agnostic local structural information. TLogic~\cite{TLogic} mines entity-independent logic rules to infer unseen entities. SST-BERT~\cite{SST-BERT} conducts inductive relation prediction by applying a time masking MLM task to pre-train BERT with structured sentences. FILT~\cite{FILT} adopts a meta-learning-based model with entity concepts to handle unseen entities in TKGs.

\noindent \textbf{Prefix Tuning.}
Prompts are manually designed textual templates to query a language model, and they are beneficial to help language models solve different tasks with all parameters frozen. To alleviate the suboptimal performance caused by discrete prompting, continuous prompts with trainable embeddings are added to the embeddings of input sequence \cite{P-tuning, Prompt_tuning}, which have been shown to achieve competitive performance across various NLP tasks. \citet{Prefix-tuning} adds trainable prefix vectors to each transformer layer within frozed Seq2Seq PLMs, aiming to efficiently adapt PLMs to natural generation tasks. \citet{CS-PromKG} introduces conditional soft prompts to sufficiently incorporate textual semantics into structural information for KGC tasks. \citet{Edgeformers} integrates local structure information into transformer layer text encoding via virtual node tokens.

\section{Method}

\begin{figure*}
    \centering
    \includegraphics[width=1\linewidth]{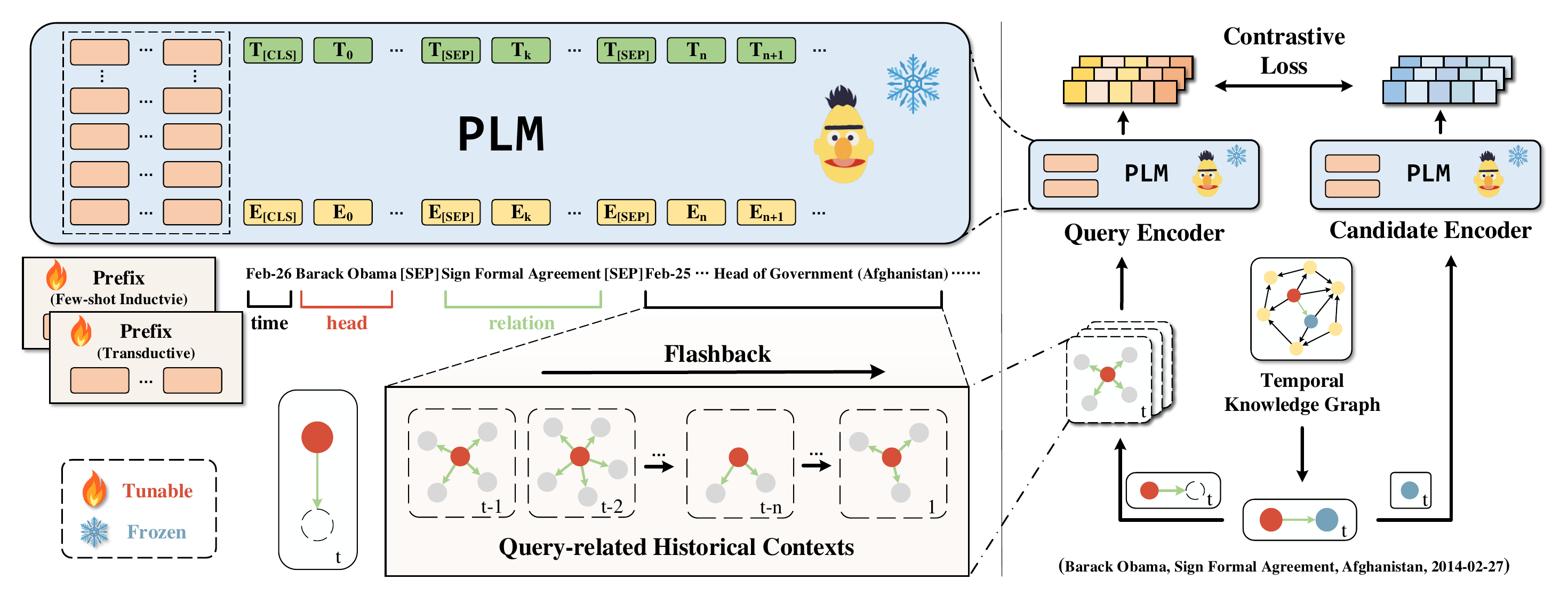}
    \caption{Overall illustration of the ChapTER model: 1) an example of ChapTER for the TKG input query (\textit{Barack Obama, Sign Formal Agreement, ?, 2014-02-07}) and corresponding candidate \textit{Afganistan} (right); 2) a detailed sketch about the structure and verbalized input of Query Encoder (left).}
    \label{fig:Model Framework}
\end{figure*}

In this section, we first give out the preliminaries and formulation of \textit{Temporal Knowledge Graph Reasoning} in Sec.\ref{sec:preliminaries}. Then we introduce the detail model framework from Sec.\ref{sec:model framework} to Sec.\ref{sec:training and inference}.

\subsection{Preliminaries} \label{sec:preliminaries}
\noindent \textbf{Temporal Knowledge Graph (TKG).} TKG is a directed graph with a collection of fact quadruples. Let $\mathcal{G}$ = $\{\mathcal{E, R, T, F}\}$ be a TKG instance, where $\mathcal{E}$, $\mathcal{R}$, $\mathcal{T}$ represent the set of entities, relations and timestamps respectively. $\mathcal{F}$ denotes the set of quadruples $(s, p, o, t)$, in which $s \in \mathcal{E}$ is a subject (head) entity, $o \in \mathcal{E}$ is an object (tail) entity, and $r \in \mathcal{R}$ is the predicate (relation) appearing at time $t$ between $s$ and $o$.  Under this definition, a TKG can be represented as a sequence of KGs $\{\mathcal{E, R, F}_{t}\}$, where $\mathcal{F}_t$ is the set of facts that occurred at time $t$.

\noindent \textbf{Transductive TKG Reasoning.}
The TKG reasoning task under transductive setting aims to answer the queries including $(s, p, ?, t_q)$ and $(?, p, o, t_q)$. Following the extrapolation setting~\cite{CyGNet}, training, validation and test sets are KGs from timestamps $T_0$ to $T_1$, $T_1$ to $T_2$, $T_2$ to $T_3$ ($ T_0 < T_1 < T_2 < T_3$).

\noindent \textbf{Few-shot Inductive TKG Reasoning.}
Under the inductive setting of TKG reasoning, given an observed background TKG $\mathcal{G}_{B} \subseteq \mathcal{E}_B \times \mathcal{R} \times \mathcal{E}_B \times \mathcal{F}$, unseen entity $e'$ is a fact from the set $\mathcal{E}'$, where $\mathcal{E}_B \cap \mathcal{E}' = \emptyset$. Hypothesizing that there are $K$ observed associated quadruple facts for each unseen entity $e'$, denoting as $(e', p, e^*, t)$ or $(e^*, p, e', t)$, where $e^* \in \mathcal{E_B} \cup \mathcal{E}'$. The goal of inductive few-shot TKGC reasoning is to answer the queries like $(e', p, ?, t_q)$ or $(?, p, e', t_q)$ from unobserved quadruples with unseen entities.

\subsection{The ChapTER Model Framework} \label{sec:model framework}
By converting the TKG reasoning problem into a query-candidate matching problem, the goal of ChapTER Model is to model the historical information and balance them with the textual semantics appropriately via a contrastive manner.

As illustrated in Figure~\ref{fig:Model Framework}, ChapTER is a pseudo-Siamese network consisting of two encoders: the query encoder $\mathcal{M}_q$ and the candidate encoder $\mathcal{M}_k$. To encode the textual information from entities and relations, we adopt the transformer-based PLM as our encoder, e.g. BERT~\cite{BERT} and RoBERTa~\cite{RoBERTa}. $\mathcal{M}_q$ and $\mathcal{M}_k$ are initialized with the same weight but tuned with prefix prompts separately. Given the query $q = (s, p, ?, t_q)$ under the tail prediction setting, $\mathcal{M}_q$ is the query encoder that aims to obtain the time-conditional entity-relation embedding $\boldsymbol{h}_q$ containing textual information of $(s,p)$ pair with historical contexts constrained on time $t$. Similarly, the candidate encoder $\mathcal{M}_k$ encodes the textual embedding $\boldsymbol{h}_k$ of candidate entity $o$. We take the mean pooling of the last-layer hidden state from $\mathcal{M}_q$ and $\mathcal{M}_k$ as the embeddings of $\boldsymbol{h}_q$ and $\boldsymbol{h}_k$:
\begin{equation}
\begin{split}
\boldsymbol{h}_q = \mathcal{M}_q({\cal P}(q)), \ \boldsymbol{h}_k = \mathcal{M}_k({\cal P}(k)),
\end{split}
\end{equation}
where $\mathcal{P}(q)$ and $\mathcal{P}(k)$ denote the input text of query and candidate separately.

% Contrastive Estimation
With the obtained embeddings $\boldsymbol{h}_q$ and $\boldsymbol{h}_k$, the score of a quadruple $(s, p, o, t)$ can be regarded as the cosine similarity between $\boldsymbol{h}_q$ and $\boldsymbol{h}_k$ by simply performing a dot-product of these two embeddings:
\begin{equation}
\begin{split}
f(s,p,o,t)\!=\!\cos (\boldsymbol{h}_q, \boldsymbol{h}_k)\!=\!\frac{\boldsymbol{h}_q \cdot \boldsymbol{h}_k}{\left \| \boldsymbol{h}_q \right \|\left \| \boldsymbol{h}_k \right \|  }.
\end{split}
\end{equation}
% Contrastive Estimation
Hence, for the query $q$, we compute the cosine similarity between query embedding $\boldsymbol{h}_q$ and all other candidate entity embeddings $\boldsymbol{h}_k$, and take the one with the highest score as the final prediction:
\begin{equation}
\begin{split}
\mathop{\arg\max} \limits_{k_i} \cos (\boldsymbol{h}_q, \boldsymbol{h}_{k_i}), \ k_i\!=\!o_i \in \mathcal{E}.
\end{split}
\end{equation}

\subsection{Text Representation \& Verbalization} \label{sec:Verbalization}

\noindent \textbf{Fact Description.}
To represent the entity or relation in a quadruple, the most representative information is the name text, e.g. "\textit{Zalmai Rassoul}" for entity and "\textit{Make statement}" for relation in dataset ICEWS~(Integrated Crisis Early Warning System)~\cite{ICEWS}. Despite this, the name texts often turn out to be short and highly overlapped, which may refer to multiple entities in the training corpus of PLM models and lead to the problem of ambiguity. To avoid this problem, we enrich the expression of semantic information by introducing the description text. More concretely, we formulate entity descriptions of ICEWS by including the hierarchical text from \textit{Country} field and \textit{Sector} field. For example, the description of entity "\textit{Virtue Party}" is "\textit{Turkey, Sunni International Religious}". Based on this, we concatenate the name and description text together as the complete entity description $D_s$. As for relation, we directly use its name text $D_r$.

\noindent \textbf{Verbalization with Historical Context.}
Given a candidate quadruple fact $(s, p, o, t)$ under the tail entity prediction task, we divide it into two parts: the time-conditional entity-relation query $q = (s, p, ?, t_q)$ and the candidate entity $k = o$. For the query $q$, we hypothesize that the historical information remains in the neighbor pairs of entity-relation-related context, which contains both $s$ and $p$ of query $q$. Specifically, we define the set of historical quadruples as follows:
\begin{equation}
\begin{split}
\mathcal{H}(s, p, t) = \big \{(s, p, \tilde{o}, t')\ |\ (s, p, \tilde{o}, t') \in {\cal F}, \\ \tilde{o} \neq o, t' \le t \big \}.
\end{split}
\end{equation}
With historical context of the quadruple fact, we individually represent the query $q$ and candidate $k$ into two different prompt formats. For timestamp text $t$ in format of "yyyy-mm-dd", we replace its month number with corresponding lexical text $L_t$. Formally, we have the input ${\cal P}(q)$ of query $q$ as follows:
\begin{tcolorbox}
    [colback=gray!20, colframe=gray!100, sharp corners, leftrule={3pt}, rightrule={0pt}, toprule={0pt}, bottomrule={0pt}, left={2pt}, right={2pt}, top={5pt}, bottom={5pt}, halign=center]
$\langle cls \rangle \ L_t \ | \ D_s \ | \ D_r \ \langle sep \rangle \ \mathcal{H}(s,p,t) \ \langle sep \rangle$
\end{tcolorbox}
\noindent where $D_s$ and $D_r$ denotes the verbalized text of entity $s$ and relation $r$. Likewise, we have the input ${\cal P}(k)$ of candidate $k$ as:
\begin{tcolorbox}
    [colback=gray!20, colframe=gray!100, sharp corners, leftrule={3pt}, rightrule={0pt}, toprule={0pt}, bottomrule={0pt}, left={16pt}, right={16pt}, top={5pt}, bottom={5pt}, halign=center]
$\langle cls \rangle \ D_s \ \langle sep \rangle$
\end{tcolorbox}

\subsection{Text encoding with virtual time prefix} \label{sec:prefix-tuning}
The goal of ChapTER is to model both historical information and textual semantics from TKGs in various complex scenarios. Instead of designing a time-specific masking strategy to pre-train a model from scratch, we introduce the virtual time prefix tokens to each Transformer layer within PLM, aiming to inject both historical and semantic information into the two-tower transformer-based model encoding procedure. We apply the prefix-based tuning methods to equip our model with capabilities to handle both transductive and few-shot inductive setting tasks. Previous works~(\citealp{Prefix-tuning}; \citealp{P-tuning_v2}) have shown the effectiveness of prefix-tuning methods in facilitating models the ability to different tasks, while achieving comparable performance with only a few parameters tuned.

We employ vector $\boldsymbol{h}$ to uniformly represents $\boldsymbol{h}_q$ in $\mathcal{M}_q$ and $\boldsymbol{h}_k$ in $\mathcal{M}_k$. Denoting $\boldsymbol{h}^{(j)} \in \mathcal{R}^{n \times d}$ as the output embeddings of all tokens in input text after $j$-th ($i \ge 1$) transformer layer, we concatenate the virtual time prefix $\boldsymbol{p}$ with $m$-length token embeddings to the text token embeddings in each transformer layer as follows:
\begin{equation}
\begin{split}
\boldsymbol{\widetilde{h}}^{(j)} = \boldsymbol{p}^{(j)} \parallel \boldsymbol{h}^{(j)}, 0\le j\le L,
\end{split}
\end{equation}
where $\boldsymbol{p}^{(j)}$ indicates the virtual prefix token embeddings of $j^{th}$ layer and $\boldsymbol{\widetilde{h}}$ is the concatenated input token embeddings. Concretely, the $i^{th}$ input token of the $j^{th}$ layer is defined as:
\begin{equation}
\begin{split}
\boldsymbol{\widetilde{h}}_i^{(j)} = 
\left\{\begin{matrix}
\boldsymbol{p}_i^{(j)}, & 0 \le i < m \\[1.5mm]
\boldsymbol{e}_i^{(j)}, & (i \ge m) \wedge (j = 0) \\[1.5mm]
\operatorname{FFN}(\boldsymbol{\widetilde{h}}^{(j-1)})_i , & (i \ge m) \wedge (j \ge 1)
\end{matrix}\right.
\end{split}
\end{equation}

During the training procedure, the weight parameters of PLM models $\mathcal{M}_q$ and $\mathcal{M}_k$ are frozen and only weight parameters in prefix prompts are updated in parallel, in which we apply the multi-head attention mechanism as follows:
\begin{equation}
\begin{gathered}
\boldsymbol{\text{MHA}}(\boldsymbol{h}^{(j)}, \boldsymbol{\widetilde{h}}^{(j)}) =\mathop{\|}\limits_{i=1}^k \boldsymbol{\text{head}}_{i}(\boldsymbol{h}_{i}^{(j)}, \boldsymbol{\widetilde{h}}_{i}^{(j)}) \\
= \mathop{\|}\limits_{i=1}^k \boldsymbol{\text{softmax}}(\boldsymbol{\mathbf{Q}}^{(j)}, \boldsymbol{\widetilde{\mathbf{K}}}^{(j)\top})\boldsymbol{\widetilde{\mathbf{V}}}^{(j)},
\end{gathered}
\end{equation}
\begin{equation}
\begin{gathered}
\boldsymbol{\mathbf{Q}}^{(j)} = \boldsymbol{\mathbf{W}}_Q^{(j)}\boldsymbol{h}^{(j)}, \\ \boldsymbol{\widetilde{\mathbf{K}}}^{(j)} = \boldsymbol{\mathbf{W}}_K^{(j)}\boldsymbol{\widetilde{h}}^{(j)}, \ \boldsymbol{\widetilde{\mathbf{V}}}^{(j)} = \boldsymbol{\mathbf{W}}_V^{(j)}\boldsymbol{\widetilde{h}}^{(j)}.
\end{gathered}
\end{equation}
We keep the query ($\mathbf{Q}$) vector still but enhance the key ($\mathbf{K}$) and value ($\mathbf{V}$) vectors with prefix embeddings. By performing the asymmetric multi-head attention in each layer, prefix vectors can efficiently capture specific data characteristics in different datasets with only a few parameters tuned.

\subsection{Training and Inference} \label{sec:training and inference}
With representations of query $\boldsymbol{h_q}$ and candidate $\boldsymbol{h_k}$, in the training procedure, we apply the InfoNCE loss~\cite{InfoNCE, SMiLE} to perform contrast estimation as follows:
\begin{equation}
\begin{split}
\mathcal{L}_{cl} = -\log\frac{e^{(cos(\boldsymbol{h_q}, \boldsymbol{h_k})-\gamma)/\tau}}{e^{(cos(\boldsymbol{h_q}, \boldsymbol{h_k})-\gamma)/\tau} + \sum\limits_{i \in \mathcal{N}_{neg}} e^{cos(\boldsymbol{h_q}, \boldsymbol{h_{i}'})/\tau}},
\end{split}
\end{equation}
where $\tau$ is a learnable temperature parameter and $\gamma$ ($\gamma > 0$) is the additive margin that encourages the model to score higher for correct quadruples.

% negative samples
$\mathcal{N}_{neg}$ represents the set of negative samples during training. Instead of randomly corrupting $s$ or $p$ of existing quadruples, we formulate $\mathcal{N}_{neg}$ with three types of negative samples:
\begin{equation}
\begin{split}
\mathcal{N}_{neg} = \big \{o'\ |\ o' \in \mathcal{N}_{in} \cup \mathcal{N}_{pre} \cup \mathcal{N}_{self} \big \}.
% , (s, p, o', t) \notin {\cal F}
\end{split}
\end{equation}
Specifically, $\mathcal{N}_{in}$ represents the set of in-batch negatives, meaning that entities within the same batch can be taken as the negative sample of each other. As for pre-batch negatives $\mathcal{N}_{pre}$, we employ a dynamic queue to store entities from recent previous $k$ batches. Besides, we take head entity $s$ from tail prediction query $(s, p, ?, t_q)$ as hard self-negative $\mathcal{N}_{self}$ to diminish false predictions due to the high text overlap between query and head entity.

For inference, ChapTER first obtains the embeddings of query $(s, p, ?, t_q)$ and all candidates via $\mathcal{M}_q$ and $\mathcal{M}_k$ separately, then computes the entity ranking by the dot-product scores between them.

\section{Experimental Setup}

\noindent \textbf{Datasets.} \ We evaluate ChapTER on TKGR task in both transductive and few-shot inductive settings. For transductive TKGR, we use four widely-used event-based TKG datasets: ICEWS14, ICEWS18, ICEWS05-15~\cite{xERTE} and ICEWS14*~\cite{Hismatch}. For few-shot inductive TKGR, we use three TKG few-shot OOG benchmarks proposed in \citet{FILT}: ICEWS14-OOG, ICEWS18-OOG and ICEWS0515-OOG. For textual descriptions, existing ICEWS datasets do not provide entity description texts, so we create them by combining corresponding \textit{country} and \textit{sector} entries for each entity. Detailed dataset statistics are shown in Appendix~\ref{sec:Dataset}.

\begin{table*}[!t]\small
\renewcommand{\arraystretch}{1.1}
 \centering
 \setlength{\tabcolsep}{1.4mm}%单元格宽度
 \resizebox{\linewidth}{!}{
 \begin{tabular}{lcccccccccccc}\toprule
    \multirow{2}{*}{\textbf{Model}} & \multicolumn{3}{c}{\textbf{ICEWS14}} & \multicolumn{3}{c}{\textbf{ICEWS18}} & \multicolumn{3}{c}{\textbf{ICEWS05-15}} & \multicolumn{3}{c}{\textbf{ICEWS14*}}
    \\\cmidrule(lr){2-4}\cmidrule(lr){5-7}\cmidrule(lr){8-10}\cmidrule(lr){11-13}
             & MRR & H@3 & H@10  & MRR & H@3 & H@10  & MRR & H@3 & H@10  & MRR & H@3 & H@10\\\midrule \specialrule{0em}{1.5pt}{1.5pt}
    \textbf{\textit{Graph-Based Methods}} \\
    % Static KG methods
    DistMult~\citep{DistMult} & .162 & .179 & .253 & .102 & .103 & .213 & .287 & .322 & .475 & .154 & .172 & .239 \\
    ComplEx~\citep{ComplEx} & .213 & .231 & .352 & .210 & .235 & .399 & .317 & .357 & .520 & .325 & .361 & .507 \\
    RotatE~\citep{RotatE} & .209 & .239 & .440 & .128 & .149 & .319 & .247 & .290 & .482 & .213 & .244 & .448 \\
    % Temporal KG methods
    TTransE~\citep{TTransE} & .134 & .173 & .346 & .083 & .086 & .219 & .157 & .197 & .380 & .137 & .177 & .357 \\
    TA-DistMult~\citep{TA-DistMult-Transe} & .265 & .302 & .454 & .168 & .184 & .336 & .265 & .302 & .454 & .258 & .297 & .430 \\
    % TA-TransE~\citep{TA-DistMult-Transe} & .174 & .292 & .474 & .126 & .179 & .374 & .194 & .313 & .503 & - & - & - \\
    DE-SimplE~\citep{DE-SimplE} & .327 & .357 & .491 & .193 & .219 & .348 & .350 & .390 & .528 & .334 & .372 & .498 \\
    TNTComplEx~\citep{TNTComplEx} & .321 & .360 & .491 & .212 & .240 & .369 & .275 & .308 & .429 & .340 & .385 & .509 \\
    CyGNet~\citep{CyGNet} & .327 & .363 & .507 & .249 & .283 & .426 & .350 & .391 & .529 & .351 & .390 & .536 \\\midrule
    \textbf{\textit{PLM-Based Methods}} \\
    SimKGC~\citep{SimKGC} & .267 & .289 & .413 & .210 & .235 & .349 & \underline{.309} & .337 & .472 & .264 & .287 & .409 \\
    % GenKGC~\citep{GenKGC} & - & - & - & - & - & - & - & - & - & - & - & - \\
    KGT5~\citep{KGT5} & .261 & .297 & .453 & .221 & .250 & .396 & .264 & .295 & .411 & .217 & .238 & .351 \\
    KGT5-context~\citep{KGT5-context} & \underline{.280} & \underline{.333} & \underline{.478} & \underline{.228} & \underline{.267} & \underline{.411} & .304 & \underline{.362} & \underline{.489} & \underline{.323} & \underline{.355} & \underline{.508} \\\midrule
    \textbf{ChapTER} & \textbf{.332} & \textbf{.370} & \textbf{.515} & \textbf{.244} & \textbf{.276} & \textbf{.412} & \textbf{.331} & \textbf{.369} & \textbf{.525} & \textbf{.338} & \textbf{.380} & \textbf{.527}
    \\\bottomrule
 \end{tabular}}
 \caption{Transductive TKG reasoning performance (with time-aware metrics) on ICEWS14, ICEWS18, ICEWS05-15 and ICEWS14*. The best PLM-based method results are in \textbf{bold} and the second best results are \underline{underlined}. For fair comparison, we add corresponding timestamps of quadruples into the input text for PLM-based baselines, to equip them with the capacities of modeling time information. More results on WIKI and YAGO datasets can be found in Appendix~\ref{sec:More Comparative Study Results}.}
 \label{tab:main results transductive}
\end{table*}

\noindent \textbf{Implementation Details.} \ All experiments are carried out on 24G RTX 3090. We adpot AdamW optimizer with linear learning rate decay to train ChapTER. The query encoder and candidate encoder are initialized with parameters of \textit{bert-based-uncased}. We truncate the description token length up to 50 for entities. The learnable temperature $\tau$ is initialized to 0.05 and the additive margin is set to 0.02. We formulate pre-batch negatives $N_{pre}$ from previous 2 batches. For the settings of all baselines, we adopt their default configurations. Most of the transductive TKGR results are taken from \citet{xERTE} and few-shot inductive TKGR results are taken from \citet{FILT}. For fairness of comparison, we reimplemented SimKGC, KGT5, KGT5-context based on their open source codes to adequately incorporate temporal information. We report the metrics MRR (mean reciprocal rank) and Hits@$N$ (proportion of correct entity rank) to evaluate the performance of ChapTER. We calculate the model results under the time-aware filtered setting~\cite{Hismatch}. More detailed implementation settings can be found in Appendix~\ref{sec:Evaluation Metrics}, \ref{sec:Hyperparameters} and~\ref{sec:Implementation of PLM-based Baselines}. Codes are avaliable at this website\footnote{\url{https://github.com/GKNL/ChapTER}}.

\section{Experimental Results}
In this section, we first compare ChapTER against other competitive baselines in both transductive and few-shot inductive TKG reasoning tasks in Sec~\ref{sec:Main Results}. Then we conduct ablation study in Sec~\ref{sec:Ablation Study} to evaluate the effectiveness of each component in ChapTER. After that, we further analyze the efficiency and flexibility of ChapTER in Sec~\ref{sec:Discussion}.

\subsection{Main Results} \label{sec:Main Results}
We compared our proposed ChapTER with various competitive baselines, and the main results of transductive and few-shot inductive TKG reasoning summarized in Table~\ref{tab:main results transductive} and Table~\ref{tab:main results few-shot inductive}, respectively.

\noindent \textbf{Results on Transductive TKGR.} \ On transductive TKGR benchmarks, we compare ChapTER with both graph-based and PLM-based models. Results on four datasets show that ChapTER achieves state-of-the-art or competitive performance against baselines. Specifically, on ICEWS14 dataset, ChapTER outperforms all PLM-based methods by a substantial margin and achieves 18.5\% (from .280 to .332) relative MRR improvement. It is worth noting that ChapTER achieves better performance with only a few prefix parameters tuned compared to fully trained PLM-based baselines, which verifies the effectiveness of prefix-tuning in TKGR tasks.

% Comparing graph methods
Compared with graph-based methods, ChapTER consistently outperforms previous baselines on ICEWS14 (MRR .332 v.s. .327), and the competitive results demonstrate ChapTER holds superiority of modeling representations in future timestamps through historical contexts. We also find that ChapTER maintains modest results on ICEWS18 and ICEWS05-15. It is worth noting that events involved in these datasets are more dense and frequent with more entities, indicating that more events are happening in the same timestamp. Since CyGNet is designed to capture the facts recurrence in the appeared history, it is good at predicting events with repetitive history yet inferior in absorbing TKG texts. This explains why ChapTER marginally lags behind CyGNet on these datasets, because redundant historical events text are truncated due to the limitation of input length in PLMs.

\begin{table*}[!t]
\renewcommand{\arraystretch}{1.1}
 \centering
 \setlength{\tabcolsep}{1.8mm}%单元格宽度
 \resizebox{\linewidth}{!}{
 \begin{tabular}{lcccccccccccccccccc}\toprule
    \multirow{3}{*}{\textbf{Model}} & \multicolumn{6}{c}{\textbf{ICEWS14-OOG}} & \multicolumn{6}{c}{\textbf{ICEWS18-OOG}} & \multicolumn{6}{c}{\textbf{ICEWS0515-OOG}}
    \\\cmidrule(lr){2-7}\cmidrule(lr){8-13}\cmidrule(lr){14-19}
    & \multicolumn{2}{c}{MRR} & \multicolumn{2}{c}{H@3} & \multicolumn{2}{c}{H@10}  & \multicolumn{2}{c}{MRR} & \multicolumn{2}{c}{H@3} & \multicolumn{2}{c}{H@10}  & \multicolumn{2}{c}{MRR} & \multicolumn{2}{c}{H@3} & \multicolumn{2}{c}{H@10} \\
    \cmidrule(lr){2-3}\cmidrule(lr){4-5}\cmidrule(lr){6-7}\cmidrule(lr){8-9}\cmidrule(lr){10-11}\cmidrule(lr){12-13}\cmidrule(lr){14-15}\cmidrule(lr){16-17}\cmidrule(lr){18-19}
    & 1-S & 3-S & 1-S & 3-S & 1-S & 3-S & 1-S & 3-S & 1-S & 3-S & 1-S & 3-S & 1-S & 3-S & 1-S & 3-S & 1-S & 3-S \\\midrule
    \textbf{\textit{Inductive TKGR Methods}} \\
    ComplEx~\citep{ComplEx} & .048 & .046 & .045 & .046 & .099 & .089 & .039 & .044 & .048 & .042 & .085 & .093 & .077 & .076 & .074 & .071 & .129 & .120 \\
    BiQUE~\citep{BiQUE} & .039 & .035 & .041 & .030 & .073 & .066 & .029 & .032 & .033 & .037 & .064 & .073 & .075 & .083 & .072 & .077 & .130 & .144 \\
    TNTComplEx~\citep{TNTComplEx} & .043 & .044 & .033 & .042 & .102 & .096 & .046 & .048 & .043 & .044 & .087 & .082 & .034 & .037 & .031 & .036 & .060 & .071 \\
    TeLM~\citep{TeLM} & .032 & .035 & .021 & .023 & .063 & .077 & .049 & .019 & .045 & .013 & .084 & .054 & .080 & .072 & .077 & .072 & .138 & .151 \\
    TeRo~\citep{TeRo} & .009 & .010 & .005 & .002 & .015 & .020 & .007 & .006 & .006 & .003 & .013 & .006 & .012 & .023 & .008 & .017 & .024 & .040 \\
    MEAN~\citep{MEAN} & .035 & .144 & .032 & .145 & .082 & .339 & .016 & .101 & .012 & .114 & .043 & .283 & .019 & .148 & .017 & .175 & .052 & .384 \\
    LAN~\citep{LAN} & .168 & .199 & .199 & .255 & .421 & .500 & .077 & .127 & .067 & .165 & .199 & .344 & .171 & .182 & .180 & .191 & .367 & .467 \\
    GEN~\citep{GEN} & .231 & .234 & .250 & .284 & .378 & .389 & .171 & .216 & .189 & .252 & .289 & .351 & .268 & .322 & .308 & .362 & .413 & .507 \\
    FILT~\citep{FILT} & .278 & .321 & .305 & .357 & .410 & .475 & .191 & \underline{.266} & .209 & \textbf{.298} & .316 & .417 & .273 & \textbf{.370} & .303 & \textbf{.391} & .405 & .516 \\
    \midrule
    \textbf{\textit{PLM-Based Methods}} \\
    SimKGC~\citep{SimKGC} & \underline{.346} & \underline{.363} & \underline{.399} & \underline{.426} & \underline{.705} & \underline{.721} & \underline{.243} & .252 & \underline{.280} & .295 & \textbf{.549} & \underline{.555} & \underline{.312} & .318 & \textbf{.375} & \underline{.382} & \underline{.629} & \underline{.637} \\
    % KGT5~\citep{KGT5} & 11.16 & - & 11.05 & - & 17.91 & - & - & - & - & - & - & - & - & - & - & - & - & - \\
    \midrule
    \textbf{ChapTER} & \textbf{.364} & \textbf{.379} & \textbf{.428} & \textbf{.446} & \textbf{.750} & \textbf{.761} & \textbf{.257} & \textbf{.266} & \textbf{.284} & \underline{.296} & \underline{.547} & \textbf{.558} & \textbf{.319} & \underline{.323} & \underline{.368} & .373 & \textbf{.644} & \textbf{.648} \\
    \midrule
    \textbf{ChapTER - Zero Shot} & \multicolumn{2}{c}{.361} & \multicolumn{2}{c}{.419} & \multicolumn{2}{c}{.752}  & \multicolumn{2}{c}{.257} & \multicolumn{2}{c}{.284} & \multicolumn{2}{c}{.544}  & \multicolumn{2}{c}{.315} & \multicolumn{2}{c}{.368} & \multicolumn{2}{c}{.638} \\
    \bottomrule
 \end{tabular}}
 \caption{Few-shot inductive TKG reasoning performance on ICEWS14-OOG, ICEWS18-OOG and ICEWS0515-OOG. The best results are in \textbf{bold} and the second best results are \underline{underlined}.}
 \label{tab:main results few-shot inductive}
\end{table*}

\noindent \textbf{Results on Few-shot Inductive TKGR.} \ As shown in Table~\ref{tab:main results transductive}, we verify ChapTER's TKG reasoning performance in a more complex few-shot inductive scenario on ICEWS14-OOG, ICEWS18-OOG and ICEWS0515-OOG datasets, considering both 1-shot and 3-shot settings. It can be seen that ChapTER substantially outperforms existing few-shot inductive TKGR methods. Concretely, ChapTER achieves striking improvement in hit@10 on ICEWS14-OOG (from .410 to .750 in 1-shot, from .475 to .761 in 3-shot), though being slightly worse on Hit@3 (3-shot) than FILT.
% Zero shot performance
We also report the zero-shot performance of ChapTER on these three datasets, and we can observe that ChapTER consistently outperforms all baselines, though slightly lags behind on few-shot performances. The overall remarkable performance verifies that ChapTER can transfer knowledge from known training entities to unseen ones, with prior knowledge and encoded historical information in PLMs.

\begin{table}[!t]
 \centering
 \setlength{\tabcolsep}{1.4mm}%单元格宽度
 \resizebox{\linewidth}{!}{
 \begin{tabular}{clcccc}\toprule
    \multirow{2}{*}{\textbf{No.}} & \multirow{2}{*}{\textbf{Model}} & \multicolumn{2}{c}{\textbf{ICEWS14}} & \multicolumn{2}{c}{\textbf{ICEWS14-OOG}} \\
    \cmidrule(lr){3-4}\cmidrule(lr){5-6}
    & & MRR & H@10  & MRR & H@10\\
    \midrule
    1 & ChapTER & \textbf{.332} & \textbf{.515} & \textbf{.361} & .752 \\
    2 & \textit{\ w/o} timestamp & .321 & .505 & .358 & .739 \\
    3 & \textit{\ w/o} description & .319 & .497 & .308 & .651 \\
    4 & \textit{\ w/o} historical contexts & .324 & .484 & .350 & .706 \\
    \midrule
    \specialrule{0em}{1pt}{1pt}\midrule
    5 & \textit{\ w/o} pre-batch neg & .326 & .509 & .350 & .726 \\
    6 & \textit{\ w/o} self neg & .329 & .510 & .358 & \textbf{.755} \\
    7 & \textit{\ w/o} pre-batch \& self neg & .326 & .507 & .343 & .734 \\
    \bottomrule
 \end{tabular}}
 \caption{Ablation study of components in ChapTER on ICEWS14 and ICEWS14-OOG (zero-shot). Lines 2-4 report variants on input prompt composition, lines 5-7 report variants on negatives combination.}
 \label{tab:Ablation Study}
\end{table}

\subsection{Ablation Studies} \label{sec:Ablation Study}
To further analyze how each component of ChapTER contributes to the final performance, we conduct ablation studies on ICEWS14 and ICEWS14-OOG, and complete results are reported in Table~\ref{tab:Ablation Study}. More ablation study results on training strategies and prompt length can be found in Appendix~\ref{sec:More Ablation Results}.

\noindent \textbf{Input Text Composition.} To verify the effectiveness of text verbalization approach mentioned in Sec.~\ref{sec:Verbalization}, we consider three variants by selectively removing the timestamp text, description text and historical contexts separately. In Table~\ref{tab:Ablation Study}, lines 2-4 show the performance of ChapTER with different input compositions. Compared to ChapTER, the performance of each ablated variant exhibits marginal decreases. Intuitively, removing description texts produces the largest performance drop (MRR drops 3.9\% and 14.7\% separately), since PLM-based models fundamentally rely on text quality. With more informative entity descriptions, the performance of ChapTER can be further improved. Moreover, lines 3-4 support the importance of temporal historical information for ChapTER. We argue this phenomenon for two reasons: 1) Though timestamps contain essential temporal information, they tend to be terse and drowned out by verbose text; 2) In contrast, historical contexts contribute more substantially to temporal modeling, as they introduce sequenced, up-to-date histories that provide abundant background for queries. Thus, ChapTER models can capture more accurate temporal information from rich historical contexts.

\begin{figure}[!t]
    \centering
    \includegraphics[width=3in]{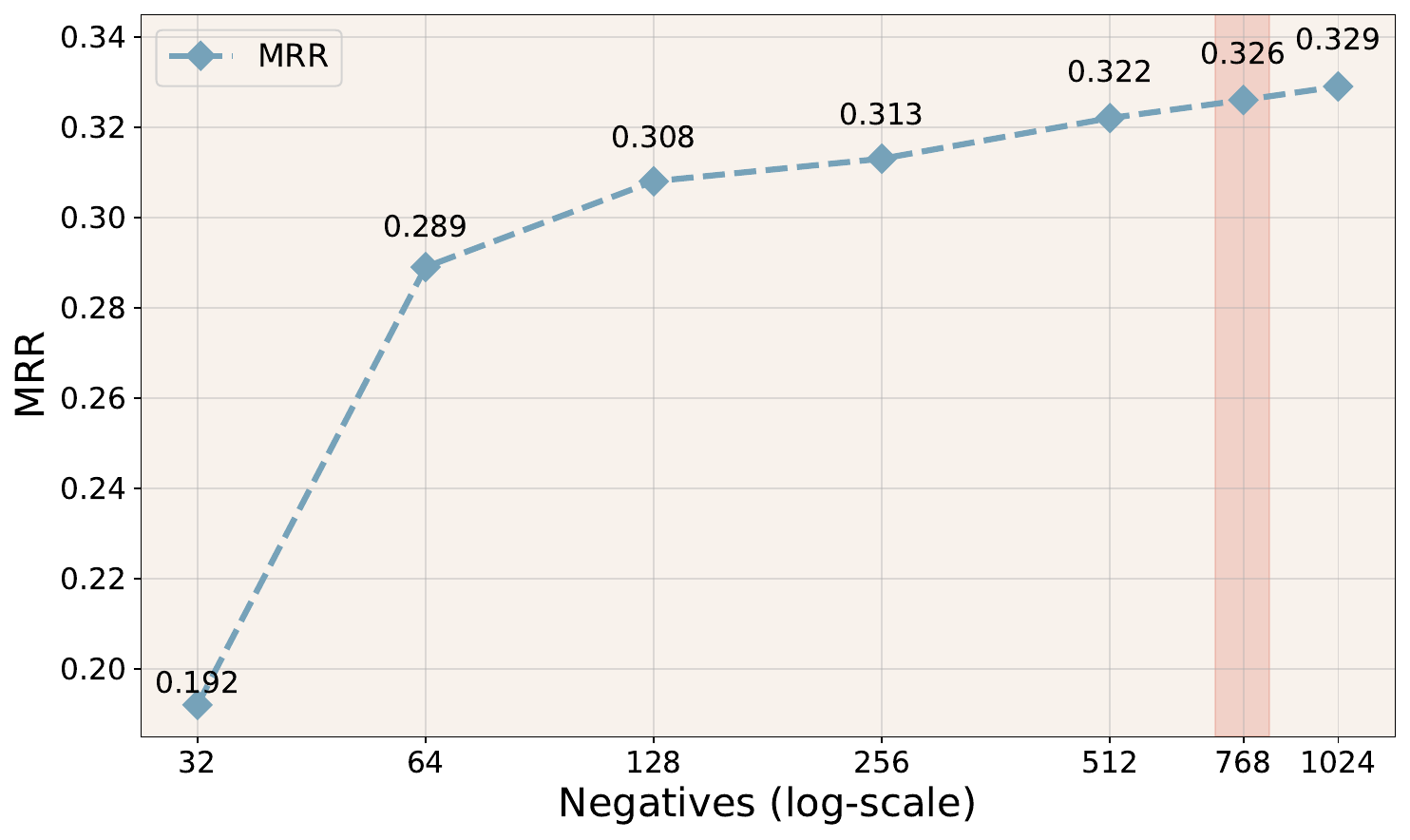}
    \caption{MRR results on ICEWS14 with in-batch negatives number changing in ChapTER.}
    \label{fig:Nagative Batchsize}
\end{figure}

\noindent \textbf{Negative Sample Combination.}
Table~\ref{tab:Ablation Study} Lines 5-7 show the performance of ChapTER with different training negatives (in-batch $\mathcal{N}_{in}$, pre-batch $\mathcal{N}_{pre}$ and self negatives $\mathcal{N}_{self}$) on ICEWS14 and ICEWS14-OOG. Three ablated variants were evaluated by separately removing $\mathcal{N}_{pre}$, $\mathcal{N}_{self}$, or both from ChapTER. We observe that removing both $\mathcal{N}_{pre}$ and $\mathcal{N}_{self}$ yields worse empirical results than removing them separately. It's worth noting that since self-negatives diminish the rely of ChapTER on naive text match, they tend to improve Hits@1 but hurt Hits@10 (e.g. Hits@10 from .755 to .752 on ICEWS14-OOG). Furthermore, we investigate the impact of in-batch negative sample numbers during model training, as shown in Figure~\ref{fig:Nagative Batchsize}. By increasing the number of negative samples, there is a steady improvement from .192 to .322, but it only obtains slight change when the number is larger than 768 (red bar). In summary, each kind of negative contributes to the best results of ChapTER.

\subsection{Discussion} \label{sec:Discussion}
In this section, we conduct further analysis on model efficiency, the ability to capture temporal information, the impact of different tuning strategies and the impact of different PLM models.

\vspace{1ex} % 段间距
\noindent \textbf{Q1: How efficient ChapTER is compared to other PLM-based models?}
\begin{table}[!t] \small
\renewcommand{\arraystretch}{0.9}
 \centering
 \setlength{\tabcolsep}{1.4mm}%单元格宽度
 \resizebox{\linewidth}{!}{
 \begin{tabular}{llccc}\toprule
    Model & PLM & Total & Trainable & T$_{train}$/ep \\
    \midrule
    \multirow{2}{*}{SimKGC} & Bert-base & 218.9M & 218.9M & 3.2min \\
    & Bert-large & 670.3M & 670.3M & 20.0min \\
    \midrule
    \multirow{2}{*}{KGT5} & T5-small & 60.5M & 60.5M & 27.6min \\
    & T5-base & 223M & 223M & 42.6min \\
    \midrule
    \multirow{2}{*}{KGT5-context} & T5-small & 60.5M & 60.5M & 38.5min \\
    & T5-base & 223M & 223M & 55.5min \\
    \midrule
    \specialrule{0em}{1pt}{1pt}\midrule
    \multirow{2}{*}{ChapTER} & Bert-base & 219.3M & 0.37M & 3.3min \\
    & Bert-large & 671.3M & 0.99M & 14.0min \\
    \bottomrule
 \end{tabular}}
 \caption{Model Efficiency of ChapTER on ICEWS14 comparing to other PLM-based methods. \textit{Total} and \textit{Trainable} indicates the total and trainable parameters.}
 \label{tab:Model Efficiency}
\end{table}
Table~\ref{tab:Model Efficiency} summarizes the model efficiency of ChapTER compared to other PLM-based methods. By taking advantage of the efficient tuning strategy, ChapTER achieves superior performance with minimal parameters tuned and reduced training time. Compared with SimKGC, ChapTER is 1.4x time faster in training with only 0.15\% parameters tuned (0.99M v.s. 760.3M). Considering recently proposed sequence-to-sequence KGR models, ChapTER outperforms KGT5 with 0.6\% parameters trained and 12x faster training time, this is because KGT5 needs to train a T5 model from scratch with task-specific input prompts. Besides, during inference, KGT5 is computationally expensive (0.83min v.s. 95.23min) due to a huge decoding search space. This suggests that ChapTER is more efficient in time and computation while achieving superior performance.

\vspace{1ex} % 段间距
\noindent \textbf{Q2: How does ChapTER use the temporal history information of events?} 
We further investigate how ChapTER actually utilizes the historical context information. As shown in Table~\ref{tab:Discuss on history modeling}, we analyze the impact of history modeling in three aspects: "Timestamp Text", "Historical Sequence Order" and "Form of Context". We can observe that removing timestamps or using random timestamps in text input both lead to a performance drop. As for historical sequence, we find that model with history in a descending order performs better than those with ascending or random order. It evidences that recent historical events are more decisive to future forecasting. Besides, we formalize the historical contexts in two ways: \textit{Entity} (e.g., all historical entities that are related to query) and \textit{Pair} (e.g., a list of complete historical quadruples). Results show that concatenate contexts by pairs achieve a higher performance than entities. We believe this is because paired contexts provide more concrete and sequenced event history. In summary, ChapTER is capable of modeling temporal information from recent and complete historical contexts.

\begin{table}[!t]
\renewcommand{\arraystretch}{0.9}
 \centering
 \setlength{\tabcolsep}{1.6mm}%单元格宽度
 \resizebox{\linewidth}{!}{
 \begin{tabular}{lcccc}\toprule
    \multirow{2}{*}{\textbf{Model}} & \multicolumn{2}{c}{\textbf{ICEWS14}} & \multicolumn{2}{c}{\textbf{ICEWS14-OOG}} \\
    \cmidrule(lr){2-3}\cmidrule(lr){4-5}
    & MRR & H@10  & MRR & H@10\\
    \midrule
    ChapTER & \textbf{.332} & \textbf{.515} & \textbf{.361} & \textbf{.752} \\
    \midrule
    \textbf{\textit{Timestamp Text}} &  &  &  & \\
    \ w/o timestamp & .321 & .505 & .358 & .739 \\
    \ random timestamp & .319 & .497 & .355 & .732 \\
    \midrule
    \textbf{\textit{Historical Sequence}} & & & & \\
    \ history descending order & .332 & .515 & .361 & .752 \\
    \ history ascending order & .322 & .498 & .344 & .721 \\
    \ history random order & .328 & .504 & .349 & .743 \\
    \midrule
    \textbf{\textit{Form of Context}} & & & & \\
    \ pairs & .332 & .515 & .361 & .752 \\
    \ entities & .324 & .502 & .359 & .734 \\
    \bottomrule
 \end{tabular}}
 \caption{Performance of different historical modeling approaches on ICEWS14 and ICEWS14-OOG datasets.}
 \label{tab:Discuss on history modeling}
\end{table}

\vspace{1ex} % 段间距
\noindent \textbf{Q3: How do different tuning strategies affect ChapTER's performance?} As mentioned in Sec.\ref{sec:prefix-tuning}, ChapTER is tuned with virtual prefix prompts with the PLM parameters frozen. To further discuss the impact of different prefix tuning methods, we compare two widely used approaches: Prefix-tuning~\cite{Prefix-tuning} and P-tuning V2~\cite{P-tuning_v2}. As summarized in Table~\ref{tab:Discuss on prefix-tuning methods}, we can observe ChapTER (P-tuning V2 with MLP reparameterization encoder) achieves better performance than the one with prefix tuning (MRR .332 v.s. .321). We also evaluate the impact of reparameterization module on ChapTER with P-tuning v2. The result show that more parameters in MLP bring a marginally improvement on performance, but its effect is inconsistent across datasets and task settings.

\begin{table}[!t]
 \centering
 \setlength{\tabcolsep}{1.4mm}%单元格宽度
 \resizebox{\linewidth}{!}{
 \begin{tabular}{llccccc}\toprule
    Method & Re-param & N$_{param}$ & T$_{train}$/ep & MRR & H@3 & H@10 \\
    \midrule
    ChapTER & \multirow{2}{*}{MLP} & \multirow{2}{*}{47.3M} & \multirow{2}{*}{4.2min} & \multirow{2}{*}{.321} & \multirow{2}{*}{.358} & \multirow{2}{*}{.503} \\
     - Prefix Tuning &  &  &  &  & \\
    \midrule
    ChapTER & \multirow{2}{*}{MLP} & \multirow{2}{*}{19.7M} & \multirow{2}{*}{3.6min} & \multirow{2}{*}{\textbf{.332}} & \multirow{2}{*}{\textbf{.370}} & \multirow{2}{*}{\textbf{.515}} \\
     - P-tuning v2 &  &  &  &  &  \\
    \midrule
    ChapTER & \multirow{2}{*}{Embedding} & \multirow{2}{*}{0.37M} & \multirow{2}{*}{3.3min} & \multirow{2}{*}{.308} & \multirow{2}{*}{.345} & \multirow{2}{*}{.491} \\
     - P-tuning v2 &  &  &  &  & \\
    \bottomrule
 \end{tabular}}
 \caption{ICEWS14 results of ChapTER with different prefix tuning methods. \textit{Re-param} denotes the reparameterization encoder and \textit{Num-param} denotes the corresponding trainable parameter numbers (on Bert-base-uncased).}
 \label{tab:Discuss on prefix-tuning methods}
\end{table}

\vspace{1ex} % 段间距
\noindent \textbf{Q4: How do PLM models affect ChapTER's performance?}
\begin{figure}[!t]
    \centering
    \subfloat[ICEWS14]{\includegraphics[width=1.5in]{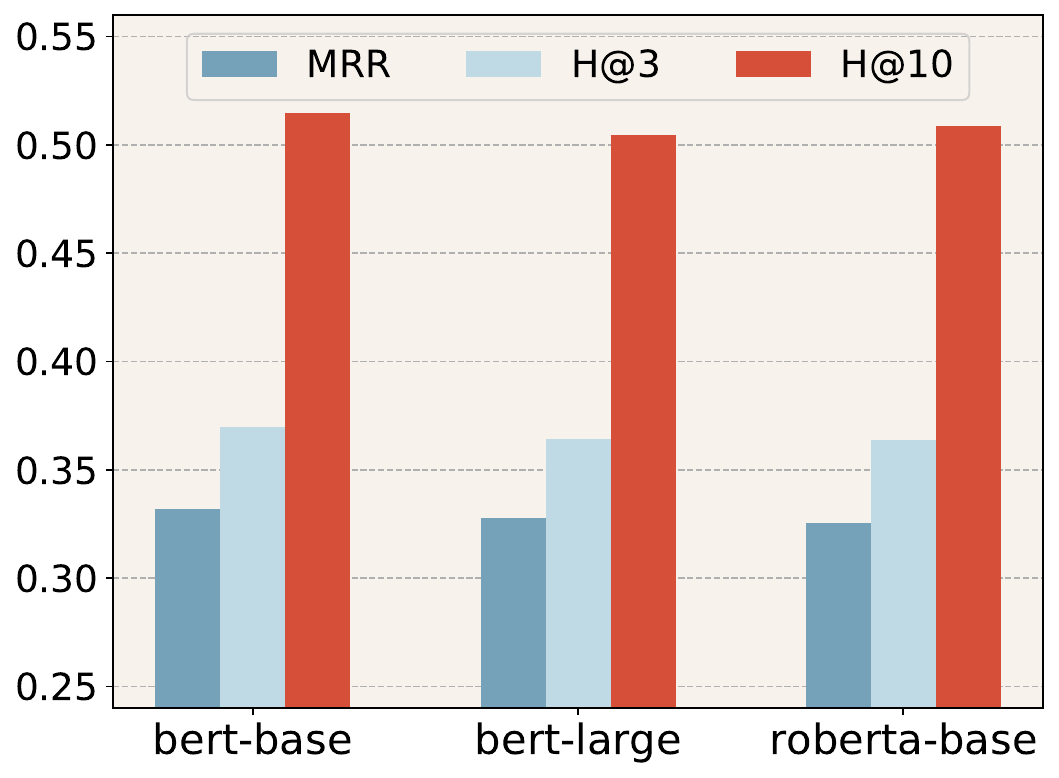} \label{PLM_variant_ICE14}} % 2.25
    % \hfill
    \subfloat[ICEWS14-OOG]{\includegraphics[width=1.5in]{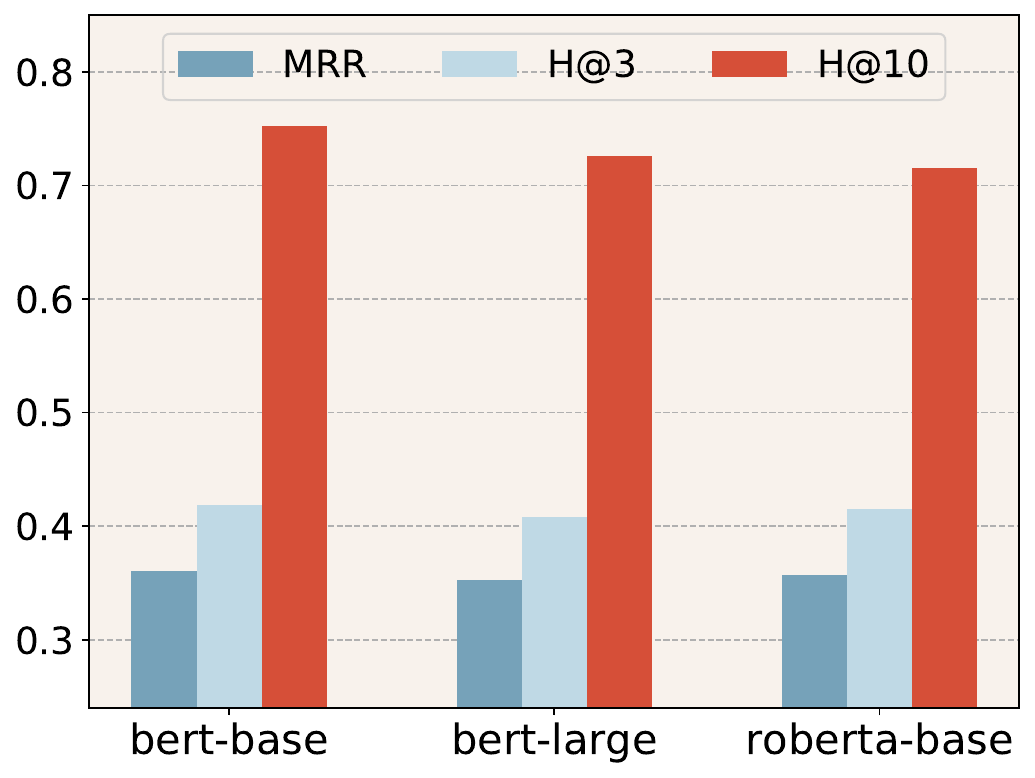} \label{PLM_variant_ICE14_OOG}} % 2.25
    \caption{Comparison of ChapTER with different PLM models on ICEWS14 and ICEWS14-OOG datasets.}
    \label{fig:Discuss on different PLMs}
\end{figure}
Figure~\ref{fig:Discuss on different PLMs} compares the model performance of ChapTER with different PLM models. We can observe that ChapTER with three PLM models all achieve close high performance on two datasets, and the utilization of Bert-base yields a marginally better result. This result suggests that our ChapTER is is robust to different PLMs with varying parameters and agnostic to PLM size. Beyond this, it is possible to improve the performance of ChapTER with some PLMs that have longer input contexts (e.g., longformer), and we leave such extensions for future studies.

\section{Conclusion and Future Work}
In this paper, we propose ChapTER, a PLM-based pseudo-Siamese framework that models balanced textual knowledge and historical information. With the introduced time prefix tokens, ChapTER is capable for TKG reasoning tasks in various complex scenarios through prefix-based tuning. Experimental results on two TKGR tasks demonstrate the superiority of ChapTER compared to competitive baselines. Thorough analysis shows the efficiency and flexibility of ChapTER. In the future, we would like to explore 1) bridging the gap of two towers with shared time prefix tuning; 2) extending our method to Seq2Seq PLMs to model temporal knowledge in a generative manner.

\section*{Limitations}
ChapTER is able to balance the textual knowledge and temporal information for the TKGR tasks in various scenarios. However, 1) ChapTER is based on PLMs and it relies on unstructured texts like entity names and descriptions. Thus the performance of ChapTER can be affected due to the quality of texts, and it could be further improved on datasets with more informative texts. Compared to ICEWS datasets, in which we manually construct description texts by concatenating the \textit{Country} and \textit{Sector} fields, datasets like Wikidata containing more informative text descriptions may result in better. 2) Due to the essence of virtual tokens in prefix tuning, which contain few parameters to be tuned compared to frozen PLMs, it may cause a collapse on tiny datasets with sparse quadruples and entities. Besides, an appropriate choice of prefix length and learning rate is crucial. We plan to work on these issues in the future work.

% \section*{Ethics Statement}

\section*{Acknowledgements}
We would like to thank all the anonymous reviewers and area chairs for their comments. This research is supported by National Science and Technology Major Project (No.2021ZD0113304), National Natural Science Foundation of China (U23A20316), Key R\&D Project of Hubei Province (2021BAA029), General Program of  Natural Science Foundation of China (NSFC) (Grant No.62072346), and founded by Joint\&Laboratory on Credit Technology.

% Entries for the entire Anthology, followed by custom entries
\bibliography{anthology,custom}

\begin{thebibliography}{46}
\expandafter\ifx\csname natexlab\endcsname\relax\def\natexlab#1{#1}\fi

\bibitem[{Baek et~al.(2020)Baek, Lee, and Hwang}]{GEN}
Jinheon Baek, Dong~Bok Lee, and Sung~Ju Hwang. 2020.
\newblock \href {https://proceedings.neurips.cc/paper/2020/hash/0663a4ddceacb40b095eda264a85f15c-Abstract.html} {Learning to extrapolate knowledge: Transductive few-shot out-of-graph link prediction}.
\newblock In \emph{Advances in Neural Information Processing Systems 33: Annual Conference on Neural Information Processing Systems 2020, NeurIPS 2020, December 6-12, 2020, virtual}.

\bibitem[{Bordes et~al.(2013)Bordes, Usunier, Garc{\'{\i}}a{-}Dur{\'{a}}n, Weston, and Yakhnenko}]{TransE}
Antoine Bordes, Nicolas Usunier, Alberto Garc{\'{\i}}a{-}Dur{\'{a}}n, Jason Weston, and Oksana Yakhnenko. 2013.
\newblock \href {https://proceedings.neurips.cc/paper/2013/hash/1cecc7a77928ca8133fa24680a88d2f9-Abstract.html} {Translating embeddings for modeling multi-relational data}.
\newblock In \emph{Advances in Neural Information Processing Systems 26: 27th Annual Conference on Neural Information Processing Systems 2013. Proceedings of a meeting held December 5-8, 2013, Lake Tahoe, Nevada, United States}, pages 2787--2795.

\bibitem[{Boschee et~al.(2015)Boschee, Lautenschlager, O’Brien, Shellman, Starz, and Ward}]{ICEWS}
E~Boschee, J~Lautenschlager, S~O’Brien, S~Shellman, J~Starz, and M~Ward. 2015.
\newblock Integrated crisis early warning system (icews) coded event data.
\newblock \emph{URL: https://dataverse. harvard. edu/dataverse/icews}.

\bibitem[{Chen et~al.(2023{\natexlab{a}})Chen, Wang, Sun, Li, and Lam}]{CS-PromKG}
Chen Chen, Yufei Wang, Aixin Sun, Bing Li, and Kwok{-}Yan Lam. 2023{\natexlab{a}}.
\newblock \href {https://doi.org/10.18653/v1/2023.findings-acl.729} {Dipping plms sauce: Bridging structure and text for effective knowledge graph completion via conditional soft prompting}.
\newblock In \emph{Findings of the Association for Computational Linguistics: {ACL} 2023, Toronto, Canada, July 9-14, 2023}, pages 11489--11503.

\bibitem[{Chen et~al.(2023{\natexlab{b}})Chen, Xu, Su, Huang, and Dou}]{SST-BERT}
Zhongwu Chen, Chengjin Xu, Fenglong Su, Zhen Huang, and Yong Dou. 2023{\natexlab{b}}.
\newblock \href {https://doi.org/10.1145/3539618.3591700} {Incorporating structured sentences with time-enhanced {BERT} for fully-inductive temporal relation prediction}.
\newblock In \emph{Proceedings of the 46th International {ACM} {SIGIR} Conference on Research and Development in Information Retrieval, {SIGIR} 2023, Taipei, Taiwan, July 23-27, 2023}, pages 889--899.

\bibitem[{Devlin et~al.(2019)Devlin, Chang, Lee, and Toutanova}]{BERT}
Jacob Devlin, Ming{-}Wei Chang, Kenton Lee, and Kristina Toutanova. 2019.
\newblock \href {https://doi.org/10.18653/v1/n19-1423} {{BERT:} pre-training of deep bidirectional transformers for language understanding}.
\newblock In \emph{Proceedings of the 2019 Conference of the North American Chapter of the Association for Computational Linguistics: Human Language Technologies, {NAACL-HLT} 2019, Minneapolis, MN, USA, June 2-7, 2019, Volume 1 (Long and Short Papers)}, pages 4171--4186.

\bibitem[{Ding et~al.(2022)Ding, Wu, He, Ma, Han, and Tresp}]{FILT}
Zifeng Ding, Jingpei Wu, Bailan He, Yunpu Ma, Zhen Han, and Volker Tresp. 2022.
\newblock \href {https://www.akbc.ws/2022/papers/6_few_shot_inductive_learning_on} {Few-shot inductive learning on temporal knowledge graphs using concept-aware information}.
\newblock In \emph{4th Conference on Automated Knowledge Base Construction}.

\bibitem[{Garc{\'{\i}}a{-}Dur{\'{a}}n et~al.(2018)Garc{\'{\i}}a{-}Dur{\'{a}}n, Dumancic, and Niepert}]{TA-DistMult-Transe}
Alberto Garc{\'{\i}}a{-}Dur{\'{a}}n, Sebastijan Dumancic, and Mathias Niepert. 2018.
\newblock \href {https://aclanthology.org/D18-1516/} {Learning sequence encoders for temporal knowledge graph completion}.
\newblock In \emph{Proceedings of the 2018 Conference on Empirical Methods in Natural Language Processing, Brussels, Belgium, October 31 - November 4, 2018}, pages 4816--4821.

\bibitem[{Goel et~al.(2020)Goel, Kazemi, Brubaker, and Poupart}]{DE-SimplE}
Rishab Goel, Seyed~Mehran Kazemi, Marcus~A. Brubaker, and Pascal Poupart. 2020.
\newblock \href {https://doi.org/10.1609/aaai.v34i04.5815} {Diachronic embedding for temporal knowledge graph completion}.
\newblock In \emph{The Thirty-Fourth {AAAI} Conference on Artificial Intelligence, {AAAI} 2020, The Thirty-Second Innovative Applications of Artificial Intelligence Conference, {IAAI} 2020, The Tenth {AAAI} Symposium on Educational Advances in Artificial Intelligence, {EAAI} 2020, New York, NY, USA, February 7-12, 2020}, pages 3988--3995.

\bibitem[{Guo and Kok(2021)}]{BiQUE}
Jia Guo and Stanley Kok. 2021.
\newblock \href {https://doi.org/10.18653/v1/2021.emnlp-main.657} {Bique: Biquaternionic embeddings of knowledge graphs}.
\newblock In \emph{Proceedings of the 2021 Conference on Empirical Methods in Natural Language Processing, {EMNLP} 2021, Virtual Event / Punta Cana, Dominican Republic, 7-11 November, 2021}, pages 8338--8351.

\bibitem[{Hamaguchi et~al.(2017)Hamaguchi, Oiwa, Shimbo, and Matsumoto}]{MEAN}
Takuo Hamaguchi, Hidekazu Oiwa, Masashi Shimbo, and Yuji Matsumoto. 2017.
\newblock \href {https://doi.org/10.24963/IJCAI.2017/250} {Knowledge transfer for out-of-knowledge-base entities : {A} graph neural network approach}.
\newblock In \emph{Proceedings of the Twenty-Sixth International Joint Conference on Artificial Intelligence, {IJCAI} 2017, Melbourne, Australia, August 19-25, 2017}, pages 1802--1808.

\bibitem[{Han et~al.(2021)Han, Chen, Ma, and Tresp}]{xERTE}
Zhen Han, Peng Chen, Yunpu Ma, and Volker Tresp. 2021.
\newblock \href {https://openreview.net/forum?id=pGIHq1m7PU} {Explainable subgraph reasoning for forecasting on temporal knowledge graphs}.
\newblock In \emph{9th International Conference on Learning Representations, {ICLR} 2021, Virtual Event, Austria, May 3-7, 2021}.

\bibitem[{Han et~al.(2023)Han, Liao, Gu, Zhang, Ding, Gu, Koeppl, Sch{\"{u}}tze, and Tresp}]{ECOLA}
Zhen Han, Ruotong Liao, Jindong Gu, Yao Zhang, Zifeng Ding, Yujia Gu, Heinz Koeppl, Hinrich Sch{\"{u}}tze, and Volker Tresp. 2023.
\newblock \href {https://doi.org/10.18653/v1/2023.findings-acl.335} {{ECOLA:} enhancing temporal knowledge embeddings with contextualized language representations}.
\newblock In \emph{Findings of the Association for Computational Linguistics: {ACL} 2023, Toronto, Canada, July 9-14, 2023}, pages 5433--5447.

\bibitem[{Jin et~al.(2023)Jin, Zhang, Meng, and Han}]{Edgeformers}
Bowen Jin, Yu~Zhang, Yu~Meng, and Jiawei Han. 2023.
\newblock \href {https://openreview.net/pdf?id=2YQrqe4RNv} {Edgeformers: Graph-empowered transformers for representation learning on textual-edge networks}.
\newblock In \emph{The Eleventh International Conference on Learning Representations, {ICLR} 2023, Kigali, Rwanda, May 1-5, 2023}. OpenReview.net.

\bibitem[{Jin et~al.(2020)Jin, Qu, Jin, and Ren}]{RE-NET}
Woojeong Jin, Meng Qu, Xisen Jin, and Xiang Ren. 2020.
\newblock \href {https://doi.org/10.18653/v1/2020.emnlp-main.541} {Recurrent event network: Autoregressive structure inferenceover temporal knowledge graphs}.
\newblock In \emph{Proceedings of the 2020 Conference on Empirical Methods in Natural Language Processing, {EMNLP} 2020, Online, November 16-20, 2020}, pages 6669--6683. Association for Computational Linguistics.

\bibitem[{Kochsiek et~al.(2023)Kochsiek, Saxena, Nair, and Gemulla}]{KGT5-context}
Adrian Kochsiek, Apoorv Saxena, Inderjeet Nair, and Rainer Gemulla. 2023.
\newblock \href {https://doi.org/10.18653/v1/2023.repl4nlp-1.11} {Friendly neighbors: Contextualized sequence-to-sequence link prediction}.
\newblock In \emph{Proceedings of the 8th Workshop on Representation Learning for NLP, RepL4NLP@ACL 2023, Toronto, Canada, July 13, 2023}, pages 131--138.

\bibitem[{Lacroix et~al.(2020)Lacroix, Obozinski, and Usunier}]{TNTComplEx}
Timoth{\'{e}}e Lacroix, Guillaume Obozinski, and Nicolas Usunier. 2020.
\newblock \href {https://openreview.net/forum?id=rke2P1BFwS} {Tensor decompositions for temporal knowledge base completion}.
\newblock In \emph{8th International Conference on Learning Representations, {ICLR} 2020, Addis Ababa, Ethiopia, April 26-30, 2020}.

\bibitem[{Leblay and Chekol(2018{\natexlab{a}})}]{TTransE}
Julien Leblay and Melisachew~Wudage Chekol. 2018{\natexlab{a}}.
\newblock \href {https://doi.org/10.1145/3184558.3191639} {Deriving validity time in knowledge graph}.
\newblock In \emph{Companion of the The Web Conference 2018 on The Web Conference 2018, {WWW} 2018, Lyon , France, April 23-27, 2018}, pages 1771--1776.

\bibitem[{Leblay and Chekol(2018{\natexlab{b}})}]{WIKI}
Julien Leblay and Melisachew~Wudage Chekol. 2018{\natexlab{b}}.
\newblock \href {https://doi.org/10.1145/3184558.3191639} {Deriving validity time in knowledge graph}.
\newblock In \emph{Companion of the The Web Conference 2018 on The Web Conference 2018, {WWW} 2018, Lyon , France, April 23-27, 2018}, pages 1771--1776. {ACM}.

\bibitem[{Lester et~al.(2021)Lester, Al{-}Rfou, and Constant}]{Prompt_tuning}
Brian Lester, Rami Al{-}Rfou, and Noah Constant. 2021.
\newblock \href {https://doi.org/10.18653/v1/2021.emnlp-main.243} {The power of scale for parameter-efficient prompt tuning}.
\newblock In \emph{Proceedings of the 2021 Conference on Empirical Methods in Natural Language Processing, {EMNLP} 2021, Virtual Event / Punta Cana, Dominican Republic, 7-11 November, 2021}, pages 3045--3059. Association for Computational Linguistics.

\bibitem[{Li and Liang(2021)}]{Prefix-tuning}
Xiang~Lisa Li and Percy Liang. 2021.
\newblock \href {https://doi.org/10.18653/v1/2021.acl-long.353} {Prefix-tuning: Optimizing continuous prompts for generation}.
\newblock In \emph{Proceedings of the 59th Annual Meeting of the Association for Computational Linguistics and the 11th International Joint Conference on Natural Language Processing, {ACL/IJCNLP} 2021, (Volume 1: Long Papers), Virtual Event, August 1-6, 2021}, pages 4582--4597.

\bibitem[{Li et~al.(2022{\natexlab{a}})Li, Guan, Jin, Peng, Lyu, Zhu, Bai, Li, Guo, and Cheng}]{CEN}
Zixuan Li, Saiping Guan, Xiaolong Jin, Weihua Peng, Yajuan Lyu, Yong Zhu, Long Bai, Wei Li, Jiafeng Guo, and Xueqi Cheng. 2022{\natexlab{a}}.
\newblock \href {https://doi.org/10.18653/v1/2022.acl-short.32} {Complex evolutional pattern learning for temporal knowledge graph reasoning}.
\newblock In \emph{Proceedings of the 60th Annual Meeting of the Association for Computational Linguistics (Volume 2: Short Papers), {ACL} 2022, Dublin, Ireland, May 22-27, 2022}, pages 290--296.

\bibitem[{Li et~al.(2022{\natexlab{b}})Li, Hou, Guan, Jin, Peng, Bai, Lyu, Li, Guo, and Cheng}]{Hismatch}
Zixuan Li, Zhongni Hou, Saiping Guan, Xiaolong Jin, Weihua Peng, Long Bai, Yajuan Lyu, Wei Li, Jiafeng Guo, and Xueqi Cheng. 2022{\natexlab{b}}.
\newblock \href {https://doi.org/10.18653/v1/2022.findings-emnlp.542} {Hismatch: Historical structure matching based temporal knowledge graph reasoning}.
\newblock In \emph{Findings of the Association for Computational Linguistics: {EMNLP} 2022, Abu Dhabi, United Arab Emirates, December 7-11, 2022}, pages 7328--7338.

\bibitem[{Li et~al.(2021)Li, Jin, Li, Guan, Guo, Shen, Wang, and Cheng}]{RE-GCN}
Zixuan Li, Xiaolong Jin, Wei Li, Saiping Guan, Jiafeng Guo, Huawei Shen, Yuanzhuo Wang, and Xueqi Cheng. 2021.
\newblock \href {https://doi.org/10.1145/3404835.3462963} {Temporal knowledge graph reasoning based on evolutional representation learning}.
\newblock In \emph{{SIGIR} '21: The 44th International {ACM} {SIGIR} Conference on Research and Development in Information Retrieval, Virtual Event, Canada, July 11-15, 2021}, pages 408--417.

\bibitem[{Liu et~al.(2023)Liu, Peng, Xu, and Peng}]{NOODLE}
Ben Liu, Miao Peng, Wenjie Xu, and Min Peng. 2023.
\newblock \href {https://doi.org/10.1007/s11280-023-01168-w} {Neighboring relation enhanced inductive knowledge graph link prediction via meta-learning}.
\newblock \emph{World Wide Web {(WWW)}}, 26(5):2909--2930.

\bibitem[{Liu et~al.(2021{\natexlab{a}})Liu, Ji, Fu, Du, Yang, and Tang}]{P-tuning_v2}
Xiao Liu, Kaixuan Ji, Yicheng Fu, Zhengxiao Du, Zhilin Yang, and Jie Tang. 2021{\natexlab{a}}.
\newblock \href {http://arxiv.org/abs/2110.07602} {P-tuning v2: Prompt tuning can be comparable to fine-tuning universally across scales and tasks}.
\newblock \emph{CoRR}, abs/2110.07602.

\bibitem[{Liu et~al.(2021{\natexlab{b}})Liu, Zheng, Du, Ding, Qian, Yang, and Tang}]{P-tuning}
Xiao Liu, Yanan Zheng, Zhengxiao Du, Ming Ding, Yujie Qian, Zhilin Yang, and Jie Tang. 2021{\natexlab{b}}.
\newblock \href {http://arxiv.org/abs/2103.10385} {{GPT} understands, too}.
\newblock \emph{CoRR}, abs/2103.10385.

\bibitem[{Liu et~al.(2019)Liu, Ott, Goyal, Du, Joshi, Chen, Levy, Lewis, Zettlemoyer, and Stoyanov}]{RoBERTa}
Yinhan Liu, Myle Ott, Naman Goyal, Jingfei Du, Mandar Joshi, Danqi Chen, Omer Levy, Mike Lewis, Luke Zettlemoyer, and Veselin Stoyanov. 2019.
\newblock \href {http://arxiv.org/abs/1907.11692} {Roberta: {A} robustly optimized {BERT} pretraining approach}.
\newblock \emph{CoRR}, abs/1907.11692.

\bibitem[{Liu et~al.(2022)Liu, Ma, Hildebrandt, Joblin, and Tresp}]{TLogic}
Yushan Liu, Yunpu Ma, Marcel Hildebrandt, Mitchell Joblin, and Volker Tresp. 2022.
\newblock \href {https://doi.org/10.1609/aaai.v36i4.20330} {Tlogic: Temporal logical rules for explainable link forecasting on temporal knowledge graphs}.
\newblock In \emph{Thirty-Sixth {AAAI} Conference on Artificial Intelligence, {AAAI} 2022, Thirty-Fourth Conference on Innovative Applications of Artificial Intelligence, {IAAI} 2022, The Twelveth Symposium on Educational Advances in Artificial Intelligence, {EAAI} 2022 Virtual Event, February 22 - March 1, 2022}, pages 4120--4127.

\bibitem[{Mahdisoltani et~al.(2015)Mahdisoltani, Biega, and Suchanek}]{YAGO}
Farzaneh Mahdisoltani, Joanna Biega, and Fabian~M. Suchanek. 2015.
\newblock \href {http://cidrdb.org/cidr2015/Papers/CIDR15\_Paper1.pdf} {{YAGO3:} {A} knowledge base from multilingual wikipedias}.
\newblock In \emph{Seventh Biennial Conference on Innovative Data Systems Research, {CIDR} 2015, Asilomar, CA, USA, January 4-7, 2015, Online Proceedings}. www.cidrdb.org.

\bibitem[{Peng et~al.(2022)Peng, Liu, Xie, Xu, Wang, and Peng}]{SMiLE}
Miao Peng, Ben Liu, Qianqian Xie, Wenjie Xu, Hua Wang, and Min Peng. 2022.
\newblock \href {https://doi.org/10.18653/v1/2022.findings-emnlp.307} {Smile: Schema-augmented multi-level contrastive learning for knowledge graph link prediction}.
\newblock In \emph{Findings of the Association for Computational Linguistics: {EMNLP} 2022, Abu Dhabi, United Arab Emirates, December 7-11, 2022}, pages 4165--4177. Association for Computational Linguistics.

\bibitem[{Saxena et~al.(2022)Saxena, Kochsiek, and Gemulla}]{KGT5}
Apoorv Saxena, Adrian Kochsiek, and Rainer Gemulla. 2022.
\newblock \href {https://doi.org/10.18653/v1/2022.acl-long.201} {Sequence-to-sequence knowledge graph completion and question answering}.
\newblock In \emph{Proceedings of the 60th Annual Meeting of the Association for Computational Linguistics (Volume 1: Long Papers), {ACL} 2022, Dublin, Ireland, May 22-27, 2022}, pages 2814--2828.

\bibitem[{Sun et~al.(2019)Sun, Deng, Nie, and Tang}]{RotatE}
Zhiqing Sun, Zhi{-}Hong Deng, Jian{-}Yun Nie, and Jian Tang. 2019.
\newblock \href {https://openreview.net/forum?id=HkgEQnRqYQ} {Rotate: Knowledge graph embedding by relational rotation in complex space}.
\newblock In \emph{7th International Conference on Learning Representations, {ICLR} 2019, New Orleans, LA, USA, May 6-9, 2019}.

\bibitem[{Teru et~al.(2020)Teru, Denis, and Hamilton}]{GraIL}
Komal~K. Teru, Etienne~G. Denis, and William~L. Hamilton. 2020.
\newblock \href {http://proceedings.mlr.press/v119/teru20a.html} {Inductive relation prediction by subgraph reasoning}.
\newblock In \emph{Proceedings of the 37th International Conference on Machine Learning, {ICML} 2020, 13-18 July 2020, Virtual Event}, volume 119 of \emph{Proceedings of Machine Learning Research}, pages 9448--9457.

\bibitem[{Trouillon et~al.(2016)Trouillon, Welbl, Riedel, Gaussier, and Bouchard}]{ComplEx}
Th{\'{e}}o Trouillon, Johannes Welbl, Sebastian Riedel, {\'{E}}ric Gaussier, and Guillaume Bouchard. 2016.
\newblock \href {http://proceedings.mlr.press/v48/trouillon16.html} {Complex embeddings for simple link prediction}.
\newblock In \emph{Proceedings of the 33nd International Conference on Machine Learning, {ICML} 2016, New York City, NY, USA, June 19-24, 2016}, pages 2071--2080.

\bibitem[{van~den Oord et~al.(2018)van~den Oord, Li, and Vinyals}]{InfoNCE}
A{\"{a}}ron van~den Oord, Yazhe Li, and Oriol Vinyals. 2018.
\newblock \href {http://arxiv.org/abs/1807.03748} {Representation learning with contrastive predictive coding}.
\newblock \emph{CoRR}, abs/1807.03748.

\bibitem[{Wang et~al.(2022)Wang, Zhao, Wei, and Liu}]{SimKGC}
Liang Wang, Wei Zhao, Zhuoyu Wei, and Jingming Liu. 2022.
\newblock \href {https://doi.org/10.18653/v1/2022.acl-long.295} {Simkgc: Simple contrastive knowledge graph completion with pre-trained language models}.
\newblock In \emph{Proceedings of the 60th Annual Meeting of the Association for Computational Linguistics (Volume 1: Long Papers), {ACL} 2022, Dublin, Ireland, May 22-27, 2022}, pages 4281--4294.

\bibitem[{Wang et~al.(2019)Wang, Han, Li, and Pan}]{LAN}
Peifeng Wang, Jialong Han, Chenliang Li, and Rong Pan. 2019.
\newblock \href {https://doi.org/10.1609/aaai.v33i01.33017152} {Logic attention based neighborhood aggregation for inductive knowledge graph embedding}.
\newblock In \emph{The Thirty-Third {AAAI} Conference on Artificial Intelligence, {AAAI} 2019, The Thirty-First Innovative Applications of Artificial Intelligence Conference, {IAAI} 2019, The Ninth {AAAI} Symposium on Educational Advances in Artificial Intelligence, {EAAI} 2019, Honolulu, Hawaii, USA, January 27 - February 1, 2019}, pages 7152--7159. {AAAI} Press.

\bibitem[{Xu et~al.(2021)Xu, Chen, Nayyeri, and Lehmann}]{TeLM}
Chengjin Xu, Yung{-}Yu Chen, Mojtaba Nayyeri, and Jens Lehmann. 2021.
\newblock \href {https://doi.org/10.18653/v1/2021.naacl-main.202} {Temporal knowledge graph completion using a linear temporal regularizer and multivector embeddings}.
\newblock In \emph{Proceedings of the 2021 Conference of the North American Chapter of the Association for Computational Linguistics: Human Language Technologies, {NAACL-HLT} 2021, Online, June 6-11, 2021}, pages 2569--2578.

\bibitem[{Xu et~al.(2020)Xu, Nayyeri, Alkhoury, Yazdi, and Lehmann}]{TeRo}
Chengjin Xu, Mojtaba Nayyeri, Fouad Alkhoury, Hamed~Shariat Yazdi, and Jens Lehmann. 2020.
\newblock \href {https://doi.org/10.18653/v1/2020.coling-main.139} {Tero: {A} time-aware knowledge graph embedding via temporal rotation}.
\newblock In \emph{Proceedings of the 28th International Conference on Computational Linguistics, {COLING} 2020, Barcelona, Spain (Online), December 8-13, 2020}, pages 1583--1593.

\bibitem[{Xu et~al.(2023{\natexlab{a}})Xu, Liu, Peng, Jia, and Peng}]{PPT}
Wenjie Xu, Ben Liu, Miao Peng, Xu~Jia, and Min Peng. 2023{\natexlab{a}}.
\newblock \href {https://doi.org/10.18653/v1/2023.findings-acl.493} {Pre-trained language model with prompts for temporal knowledge graph completion}.
\newblock In \emph{Findings of the Association for Computational Linguistics: {ACL} 2023, Toronto, Canada, July 9-14, 2023}, pages 7790--7803.

\bibitem[{Xu et~al.(2023{\natexlab{b}})Xu, Ou, Xu, and Fu}]{CE-NET}
Yi~Xu, Junjie Ou, Hui Xu, and Luoyi Fu. 2023{\natexlab{b}}.
\newblock \href {https://doi.org/10.1609/aaai.v37i4.25601} {Temporal knowledge graph reasoning with historical contrastive learning}.
\newblock In \emph{Thirty-Seventh {AAAI} Conference on Artificial Intelligence, {AAAI} 2023, Thirty-Fifth Conference on Innovative Applications of Artificial Intelligence, {IAAI} 2023, Thirteenth Symposium on Educational Advances in Artificial Intelligence, {EAAI} 2023, Washington, DC, USA, February 7-14, 2023}, pages 4765--4773.

\bibitem[{Yang et~al.(2015)Yang, Yih, He, Gao, and Deng}]{DistMult}
Bishan Yang, Wen{-}tau Yih, Xiaodong He, Jianfeng Gao, and Li~Deng. 2015.
\newblock \href {http://arxiv.org/abs/1412.6575} {Embedding entities and relations for learning and inference in knowledge bases}.
\newblock In \emph{3rd International Conference on Learning Representations, {ICLR} 2015, San Diego, CA, USA, May 7-9, 2015, Conference Track Proceedings}.

\bibitem[{Yang et~al.(2022)Yang, Huang, Xia, and Li}]{KGCL}
Yuhao Yang, Chao Huang, Lianghao Xia, and Chenliang Li. 2022.
\newblock \href {https://doi.org/10.1145/3477495.3532009} {Knowledge graph contrastive learning for recommendation}.
\newblock In \emph{{SIGIR} '22: The 45th International {ACM} {SIGIR} Conference on Research and Development in Information Retrieval, Madrid, Spain, July 11 - 15, 2022}, pages 1434--1443.

\bibitem[{Yasunaga et~al.(2021)Yasunaga, Ren, Bosselut, Liang, and Leskovec}]{QA-GNN}
Michihiro Yasunaga, Hongyu Ren, Antoine Bosselut, Percy Liang, and Jure Leskovec. 2021.
\newblock \href {https://doi.org/10.18653/v1/2021.naacl-main.45} {{QA-GNN:} reasoning with language models and knowledge graphs for question answering}.
\newblock In \emph{Proceedings of the 2021 Conference of the North American Chapter of the Association for Computational Linguistics: Human Language Technologies, {NAACL-HLT} 2021, Online, June 6-11, 2021}, pages 535--546.

\bibitem[{Zhu et~al.(2021)Zhu, Chen, Fan, Cheng, and Zhang}]{CyGNet}
Cunchao Zhu, Muhao Chen, Changjun Fan, Guangquan Cheng, and Yan Zhang. 2021.
\newblock \href {https://doi.org/10.1609/aaai.v35i5.16604} {Learning from history: Modeling temporal knowledge graphs with sequential copy-generation networks}.
\newblock In \emph{Thirty-Fifth {AAAI} Conference on Artificial Intelligence, {AAAI} 2021, Thirty-Third Conference on Innovative Applications of Artificial Intelligence, {IAAI} 2021, The Eleventh Symposium on Educational Advances in Artificial Intelligence, {EAAI} 2021, Virtual Event, February 2-9, 2021}, pages 4732--4740.

\end{thebibliography}
\bibliographystyle{acl_natbib}

% \clearpage
\appendix

\begin{table*}[!t]\small
\renewcommand{\arraystretch}{1.2}
\centering
\setlength{\tabcolsep}{2.3mm}%单元格宽度
\begin{tabular}{l|ccccccc}
\toprule
\textbf{Dataset} & $|\mathcal{F}_{train}|$ & $|\mathcal{F}_{valid}|$ & $|\mathcal{F}_{test}|$ & $|\mathcal{E}|$ & $|\mathcal{R}|$ & Time Snapshot & Time Granularity\\
\midrule
ICEWS14 & 63,685 & 13,823 & 13,222 & 7,128 & 230 & 365 & 1 day \\
ICEWS18 & 373,018 & 45,995 & 49,545 & 23,033 & 256 & 304 & 1 day \\
ICEWS05-15 & 322,958 & 69,224 & 69,147 & 10,488 & 251 & 4,017 & 1 day \\
ICEWS14* & 74,845 & 8,514 & 7,371 & 7,128 & 230 & 365 & 1 day \\
WIKI & 539,286 & 67,538 & 63,110 & 12,554 & 24 & 232 & 1 year \\
YAGO & 51,205 & 10,973 & 10,973 & 100,38 & 10 & 194 & 1 year \\
\bottomrule
\end{tabular}
\caption{Statistics of datasets for transductive TKG reasoning. $|\mathcal{F}_{train}|$, $|\mathcal{F}_{valid}|$, $|\mathcal{F}_{test}|$ represent the number of quadruples in train sets, valid sets test sets, respectively. $|\mathcal{E}|$, $|\mathcal{R}|$ denote the number of entites and relations.}
\label{tab:transductive datasets}
\end{table*}

\begin{table*}[!t]\small
\renewcommand{\arraystretch}{1.2}
\centering
\setlength{\tabcolsep}{2.3mm}%单元格宽度
\begin{tabular}{l|cccccccc}
\toprule
\textbf{Dataset} & $|\mathcal{F}_{back}|$ & $|\mathcal{F}_{train}|$ & $|\mathcal{F}_{valid}|$ & $|\mathcal{F}_{test}|$ & $|\mathcal{E}|$ & $|\mathcal{R}|$ & Time Snapshot & Time Granularity\\
\midrule
ICEWS14-OOG & 83,448 & 5,772 & 718 & 705 & 7128 & 230 & 365 & 1 day \\
ICEWS18-OOG & 444,269 & 19,291 & 2,425 & 2,373 & 23033 & 256 & 304 & 1 day \\
ICEWS0515-OOG & 448,695 & 10,115 & 1,271 & 1,228 & 10488 & 251 & 4017 & 1 day \\
\bottomrule
\end{tabular}
\caption{Statistics of datasets for few-shot inductive TKG reasoning. $|\mathcal{F}_{train}|$, $|\mathcal{F}_{valid}|$, $|\mathcal{F}_{test}|$ represent the number of quadruples containing unseen entities in train sets, valid sets and test sets, respectively. $|\mathcal{F}_{back}|$ denotes the number of remaining quadruples without unseen entities.}
\label{tab:few-shot inductive datasets}
\end{table*}

\section{Experimental Details} \label{sec:Dataset Details}

\subsection{Dataset} \label{sec:Dataset}

We use the transductive TKGR datasets ICEWS14, ICEWS18, ICEWS05-15, YAGO from \citet{xERTE}, WIKI from \citet{WIKI} and ICEWS14* from \citet{RE-GCN}, which contain political facts of Integrated Crisis Early Warning System~\cite{ICEWS}. We take few-shot inductive datasets ICEWS14-OOG, ICEWS18-OOG and ICEWS0515-OOG from \citet{FILT}. Following the original data split, we summarize the statistics of these datasets in Table~\ref{tab:transductive datasets} and Table~\ref{tab:few-shot inductive datasets}. For the absent description texts, we find the texts of \textit{country} and \textit{sector} entries from origin data source\footnote{\url{https://dataverse.harvard.edu/dataverse/icews}} and combining them together to construct entity descriptions.

\subsection{Baselines} \label{sec:Baselines}
We compare ChapTER with several competitive state-of-the-art baselines in transductive and few-shot inductive TKG reasoning settings. For transductive TKG reasoning, we include 1) traditional KG reasoning methods, i.e. DistMult~\cite{DistMult}, ComplEx~\cite{ComplEx} and RotatE~\cite{RotatE}; 2) TKG Reasoning methods, i.e. TTransE~\cite{TTransE}, TA-DistMult~\cite{TA-DistMult-Transe}, TA-TransE~\cite{TA-DistMult-Transe}, DE-SimplE~\cite{DE-SimplE}, TNTComplEx~\cite{TNTComplEx} and CyGNet~\cite{CyGNet}; 3) PLM-based methods, i.e. SimKGC~\cite{SimKGC}, KGT5~\cite{KGT5} and KGT5-context~\cite{KGT5-context}. For few-shot inductive TKG reasoning, we include 1) traditional KGR methods BiQUE~\cite{BiQUE}; 2) traditional TKGR methods, i.e. TELM~\cite{TeLM} and TeRo~\cite{TeRo}; 3) inductive KGR methods, i.e. MEAN~\cite{MEAN} and LAN~\cite{LAN}; 4) mrta-learning-based method GEN~\cite{GEN}.

\subsection{Evaluation Metrics} \label{sec:Evaluation Metrics}
In the experiments, we report the widely used metrics MRR (Mean Reciprocal Rank) and Hits@$N$ to evaluate the performance of ChapTER under both two settings. MRR measures the average reciprocal ranks of all test triples. Hits@$N$ ($N \in$ \{1,3,10\}) calculates the proportion of correct entities ranked among the top-$N$. For fair comparison, we calculate the model results under the time-aware filtered setting~\cite{Hismatch}, which only filters out the quadruples that occur at the query time. All metrics are computed by averaging over head and tail entity prediction, and model is selected by MRR value on the validation set.

\begin{table}[!t]\small
\renewcommand{\arraystretch}{1.2}
\centering
\setlength{\tabcolsep}{1.4mm}%单元格宽度
\resizebox{\linewidth}{!}{
\begin{tabular}{l|c}
\toprule
\textbf{Hyperparameters} & \textbf{Values}\\
\midrule
Learning rate & \{1e-5, 3e-5, 5e-4, 5e-3\} \\
Prompt Length & \{2, 4, 6, 10, 15, 20, 50\} \\
In-batch negatives & \{32, 64, 128, 256, 512, 764, 1024\} \\
Queued negative batches & \{1, 2, 4\} \\
Train epoch & \{10, 15, 20\} \\
Max token number & \{50, 60, 70\} \\
\bottomrule
\end{tabular}}
\caption{Details of hyperparameters.}
\label{tab:Hyperparameters}
\end{table}

\subsection{Hyperparameters} \label{sec:Hyperparameters}
We perform grid-search on hyperparameters including learning rate, prompt length, in-batch negatives, queued negative batches, train epoch and max token number. The optimal hyperparameters are summarized in Table~\ref{tab:Hyperparameters}.

\subsection{Implementation of PLM-based Baselines} \label{sec:Implementation of PLM-based Baselines}
We reimplement SimKGC\footnote{\url{https://github.com/intfloat/simkgc}}, KGT5\footnote{\url{https://github.com/apoorvumang/kgt5}} and KGT5-context\footnote{\url{https://github.com/uma-pi1/kgt5-context}} based on their official codes. To adapt them to TKG datasets, we modify their input format from triplet to quadruplet by concatenating timestamps with their corresponding input texts. For example, a tail-prediction query input text in KGT5 can be formulated as "predict tail: 2014-01-01 | Benjamin Netanyahu | Sign formal agreement". Following their default hyperparameter setting, the hyperparameters are slightly different on different TKG datasets. For evaluation, we changed their filter setting into time-aware filter setting to align with other TKGR models.

\section{More Comparative Study Results} \label{sec:More Comparative Study Results}
We report more transductive TKGR results on WIKI~\cite{WIKI} and YAGO~\cite{YAGO} in Table~\ref{tab:main results transductive more}, since these datasets hold different distributions from ICEWS datasets.

\begin{table}[!t]\small
\renewcommand{\arraystretch}{1.2}
 \centering
 \setlength{\tabcolsep}{1.4mm}%单元格宽度
 \resizebox{\linewidth}{!}{
 \begin{tabular}{lcccccccccccc}\toprule
    \multirow{2}{*}{\textbf{Model}} & \multicolumn{4}{c}{\textbf{WIKI}} & \multicolumn{4}{c}{\textbf{YAGO}}
    \\\cmidrule(lr){2-5}\cmidrule(lr){6-9}
             & MRR & H@1 & H@3 & H@10  & MRR & H@1 & H@3 & H@10\\\midrule \specialrule{0em}{1.5pt}{1.5pt}
    \textbf{\textit{Graph-Based Methods}} \\
    % Static KG methods
    DistMult & .109 & .089 & .110 & .168 & .120 & .102 & .123 & .149 \\
    ComplEx & .245 & .197 & .273 & .348 & .121 & .104 & .124 & .148 \\
    % Temporal KG methods
    TTransE & .293 & .217 & .344 & .424 & .057 & .014 & .090 & .112 \\
    TA-DistMult & .445 & .399 & .487 & .517 & .115 & .102 & .119 & .139 \\
    DE-SimplE & .454 & .426 & .477 & .496 & .117 & .107 & .121 & .135 \\
    TNTComplEx & .450 & .400 & .493 & .520 & .120 & \underline{.111} & .121 & .136 \\
    CyGNet & .339 & .291 & .361 & .419 & .125 & .110 & .127 & .148 \\\midrule
    \textbf{\textit{PLM-Based Methods}} \\
    SimKGC & \underline{.505} & \underline{.439} & \underline{.540} & \underline{.618} & \underline{.140} & .096 & \underline{.142} & \underline{.200} \\\midrule
    \textbf{ChapTER} & \textbf{.534} & \textbf{.463} & \textbf{.570} & \textbf{.653} & \textbf{.148} & \textbf{.115} & \textbf{.144} & \textbf{.202}
    \\\bottomrule
 \end{tabular}}
 \caption{Transductive TKG reasoning performance (with time-aware metrics) on WIKI and YAGO. The best results are in \textbf{bold} and the second best results are \underline{underlined}. For fair comparison, we add corresponding timestamps of quadruples into the input text for PLM-based baselines, to equip them with the capacities of modeling time information.}
 \label{tab:main results transductive more}
\end{table}

\section{More Ablation Results} \label{sec:More Ablation Results}

\subsection{Ablation on Training Strategy} \label{sec:Ablation on Training Strategy}
We empirically evaluate the impact of different tuning layers on ChapTER. Table~\ref{tab:Ablation on Training Strategy} Lines 2-5 summarize the performance of ChapTER with different layers tuned. It can observed that ChapTER performs better than the fully tuned model, we argue that 1) ChapTER with time prefixes are effective to capture temporal information; 2) more tuning parameters may lead to an over-fitting on textual information and overlook the temporal correlations; 3) ChapTER on ICEWS14-OOG dataset is less sensitive to the changes of tuning parameters, because it is more crucial to model textual information for unseen entities in inductive TKGR. Besides, comparing line 8 and line 10,  we can find that tuning bottom layers tends to obtain a better performance than tuning top layers. This could be because lower layer in Bert can capture low-level semantic features, which is more important for TKGR tasks.

\begin{table}[!t]
 \centering
 \setlength{\tabcolsep}{1.4mm}%单元格宽度
 \resizebox{\linewidth}{!}{
 \begin{tabular}{clcccc}\toprule
    \multirow{2}{*}{\textbf{No.}} & \multirow{2}{*}{\textbf{Model}} & \multicolumn{2}{c}{\textbf{ICEWS14}} & \multicolumn{2}{c}{\textbf{ICEWS14-OOG}} \\
    \cmidrule(lr){3-4}\cmidrule(lr){5-6}
    & & MRR & H@10  & MRR & H@10\\
    \midrule
    1 & ChapTER & \textbf{.332} & \textbf{.515} & \textbf{.361} & .752 \\
    \midrule
    \specialrule{0em}{1pt}{1pt}\midrule
    2 & \textit{\ w/} last layer tuned & .323 & .504 & .347 & .748 \\
    3 & \textit{\ w/} last 6 layers tuned & .316 & .499 & .359 & .748 \\
    4 & \textit{\ w/} first layer tuned & .311 & .494 & .333 & .721 \\
    5 & \textit{\ w/} fully tuned & .316 & .493 & .357 & .748 \\
    \bottomrule
 \end{tabular}}
 \caption{Performance of ChapTER with different training strategies. Lines 2-5 report variants on training strategy.}
 \label{tab:Ablation on Training Strategy}
\end{table}

\subsection{Ablation on Prefix Length} \label{sec:Ablation on Prefix Length}
\begin{figure}[!t]
    \centering
    \subfloat[MRR \& H@3 Score]{\includegraphics[width=1.5in]{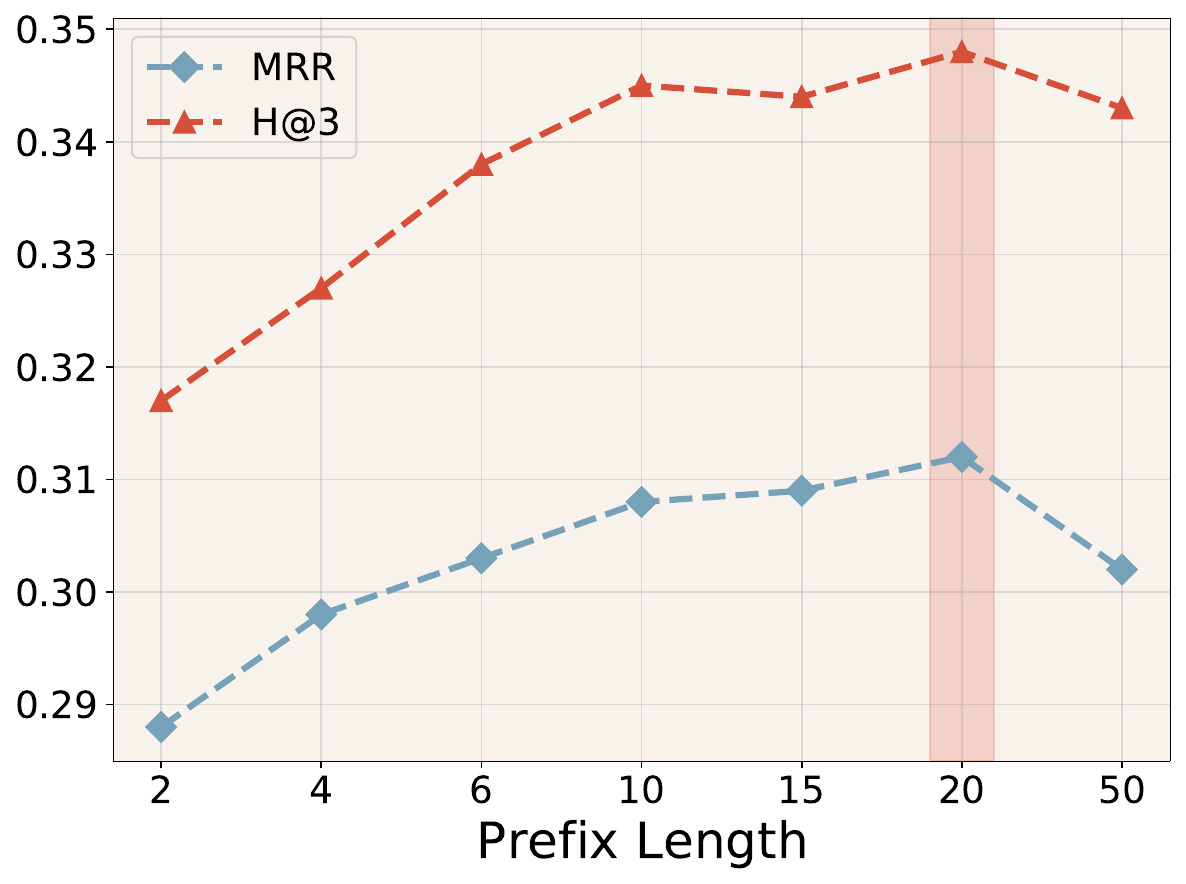} \label{prompt_length_MRR_Hit3}} % 2.25
    % \hfill
    \subfloat[Trainable Parameters]{\includegraphics[width=1.5in]{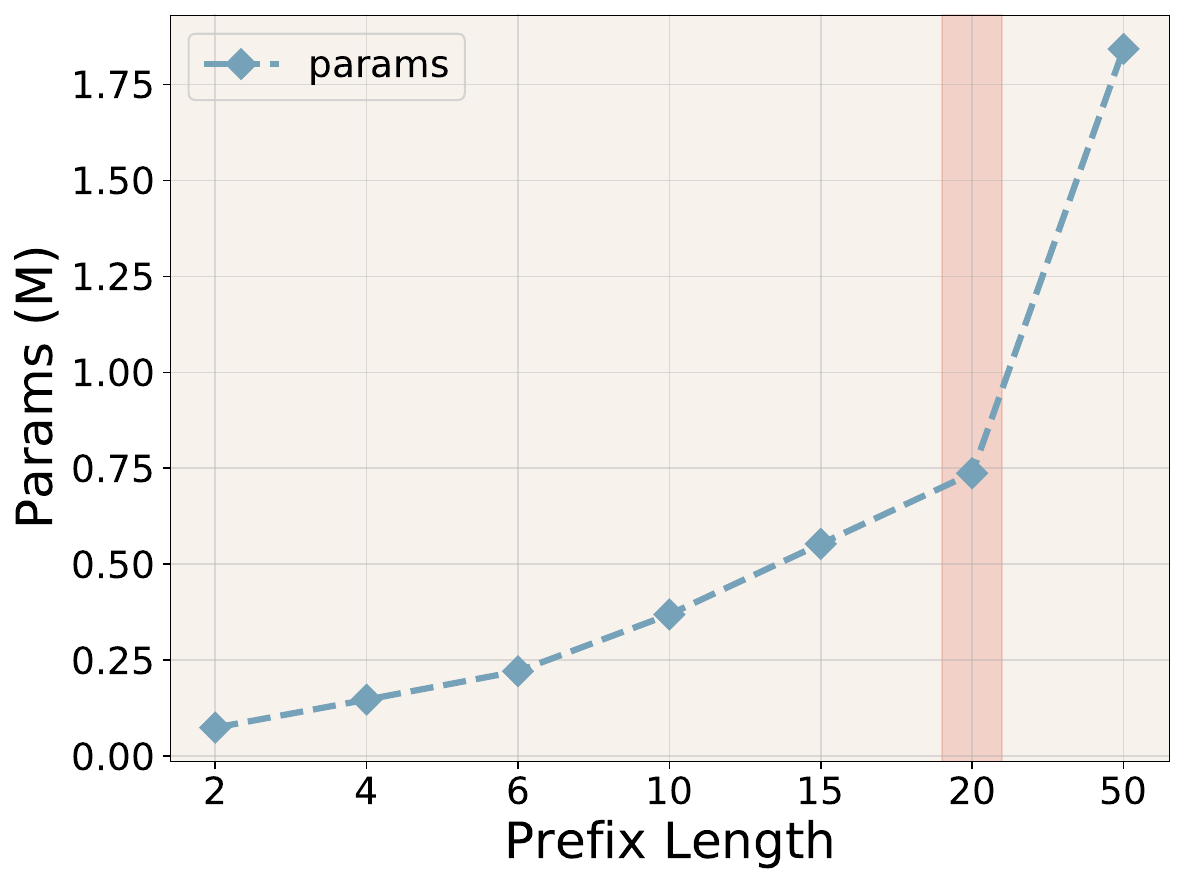} \label{prompt_length_Param}} % 2.25
    \caption{Impact (performance and parameter size) of prefix length on ICEWS14 dataset.}
    \label{fig:prefix length}
\end{figure}

We conduct extensive experiments on the impact of prefix length for ChapTER and results are shown in Figure~\ref{fig:prefix length}. As evidenced, model performance exhibits a slight positive correlation to prefix length as increasing from 2 to 20 (MRR from .288 to .312), while number of trainable parameters is also expanding (from 0.074M to 0.737M). We can also observe that a further increase of prefix length leads to an inferior performance (from .312 to .302), as additional complexity imposes considerable challenges.

\end{document}